\documentclass{IEEEoj}
\usepackage{textcomp}
\usepackage{verbatim}
\usepackage{graphicx}
\usepackage{epsf}
\usepackage{epsfig}
\usepackage{epstopdf}
\usepackage{ifpdf}
\usepackage{cite}
\usepackage[utf8]{luainputenc}
\usepackage{amssymb}
\usepackage{amsthm}
\usepackage{lipsum}

\usepackage{dsfont}

\usepackage{pifont}

\ifCLASSINFOpdf
\else
\fi

\usepackage[cmex10]{amsmath}
\usepackage{amsmath}
\DeclareMathOperator*{\argmax}{arg\,max} 
\usepackage{array}
\usepackage{mdwmath}
\usepackage{mdwtab}
\usepackage{eqparbox}
\usepackage{url}

\usepackage{breakurl}
\usepackage[breaklinks]{hyperref}
\usepackage{caption}
\usepackage{subcaption}
\usepackage{amsfonts}
\usepackage{bbm}
\usepackage{breakurl}
\usepackage{doi}
\usepackage{float}
\usepackage{epsf}
\usepackage{epsfig}
\usepackage{epstopdf}
\usepackage{algorithmic}
\usepackage[linesnumbered, ruled, vlined]{algorithm2e}
\SetKw{KwOr}{or}
\SetKw{KwAnd}{and}
\SetKw{Cnt}{continue}
\usepackage{xcolor}

\usepackage{cite}
\usepackage{amsmath,amssymb,amsfonts}
\usepackage{algorithmic}
\usepackage{graphicx,color}
\usepackage{textcomp}
\def\BibTeX{{\rm B\kern-.05em{\sc i\kern-.025em b}\kern-.08em
    T\kern-.1667em\lower.7ex\hbox{E}\kern-.125emX}}
\AtBeginDocument{\definecolor{ojcolor}{cmyk}{0.93,0.59,0.15,0.02}}
\def\OJlogo{\vspace{-4pt}\hskip-4pt\includegraphics[height=18pt]{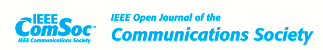}}
\begin{document}
\receiveddate{XX Month, XXXX}
\reviseddate{XX Month, XXXX}
\accepteddate{XX Month, XXXX}
\publisheddate{XX Month, XXXX}
\currentdate{11 January, 2024}
\doiinfo{OJCOMS.2024.011100}

\title{
Meta-Hierarchical Reinforcement Learning for Scalable Resource Management in O-RAN}
\author{F. Lotfi\IEEEauthorrefmark{1} \IEEEmembership{(Member, IEEE)}, F. Afghah \IEEEauthorrefmark{1}\IEEEmembership{(Senior Member, IEEE)}}
\affil{Department of Electrical \& Computer Engineering, Clemson University, Clemson, SC 29631 USA}
\corresp{CORRESPONDING AUTHOR: F. Lotfi (e-mail: flotfi@clemson.edu).}
\authornote{This material is based upon work supported by the National Science Foundation under Grant Numbers  CNS-2202972, and CNS-2232048.}
\markboth{Meta-Hierarchical Reinforcement Learning for Scalable Resource Management in O-RAN}{Lotfi \textit{et al.}}

\begin{abstract}
The increasing complexity of modern applications demands wireless networks capable of real-time adaptability and efficient resource management. The Open Radio Access Network (O-RAN) architecture, with its RAN Intelligent Controller (RIC) modules, has emerged as a pivotal solution for dynamic resource management and network slicing. While artificial intelligence (AI) driven methods have shown promise, most approaches struggle to maintain performance under unpredictable and highly dynamic conditions. 
This paper proposes an adaptive Meta-Hierarchical Reinforcement Learning (Meta-HRL) framework, inspired by Model-Agnostic Meta-Learning (MAML), to jointly optimize resource allocation and network slicing in O-RAN. The framework integrates hierarchical control with meta-learning to enable both global and local adaptation: the high-level controller allocates resources across slices, while low-level agents perform intra-slice scheduling. The adaptive meta-update mechanism weights tasks by temporal-difference error variance, improving stability and prioritizing complex network scenarios. 
Theoretical analysis establishes sublinear convergence and regret guarantees for the two-level learning process. Simulation results demonstrate a $19.8\% $improvement in network management efficiency compared with baseline RL and meta-RL approaches, along with faster adaptation and higher QoS satisfaction across eMBB, URLLC, and mMTC slices. Additional ablation and scalability studies confirm the method’s robustness, achieving up to $40\%$ faster adaptation and consistent fairness, latency, and throughput performance as network scale increases.
\end{abstract}


\maketitle

\section{INTRODUCTION}
\IEEEPARstart{N}{ext}-generation wireless networks built on the Open RAN (O-RAN) architecture are designed for flexibility, enabling operators to dynamically adapt to changing user demands and network conditions. This adaptability is primarily driven by RAN Intelligent Controller (RIC) modules, which enhance network functionality through intelligent resource management and real-time data analysis~\cite{polese2022understanding, 3gppRe18}.
These capabilities enable operators to maintain high levels of responsiveness and adaptability, ensuring that the network can effectively address diverse and evolving use cases~\cite{3gppRe18,d2022dapps}. 

Effective network configuration management ensures that modern wireless networks remain adaptable and scalable. The dynamic nature of resource allocation in O-RAN systems allows operators to modify network architecture on-the-fly, and ensures seamless operation even under fluctuating conditions. This adaptability enables networks to scale from simple configurations to complex architectures as user demands change. However, significant challenges remain in real-time resource management, particularly in unpredictable environments. These challenges are heightened in densely populated areas or hotspots, where demand surges create corner cases that challenge the network’s responsiveness. Additionally, the real-time orchestration of virtualized distributed units (DUs) and the deployment of xApps introduce further complexities, requiring advanced strategies for dynamic resource management. Existing works have explored ML-based approaches for real-time resource management~\cite{chen2023hierarchical,rezazadeh2022specialization,li2018deep,zhang2022federated,zhou2022learning,lotfi2024open,kavehmadavani2024empowering,lotfiattention,habib2023hierarchical,lee2021ran,lotfi2022evolutionary,erdol2022federated,lotfi2024meta,nagib2023accelerating,niu2023multiagent,zhang2023device,zhou2022knowledge,raftopoulos2024drl, 10071958, cheng2022reinforcement, thaliath2022predictive,basir2025}, but their applicability in dynamic and complex O-RAN environments is limited. However, existing learning-based methods often struggle with unpredictable traffic patterns, non-stationary environments, and scalability limits. 

To address these challenges, fast adaptation in O-RAN slicing and scheduling is essential for managing complexity and maintaining high-quality service. Advanced resource management strategies are indispensable for optimizing resource utilization, ensuring flexibility, and achieving rapid convergence in dynamic conditions. Building on this necessity, literature recently explored a meta-learning framework to enhance the efficiency and adaptability of Deep Reinforcement Learning (DRL) approaches, proving particularly advantageous in few-shot learning scenarios where deriving meaningful insights from small data samples is crucial~\cite{beck2023survey,yuan2021meta,erdol2022federated,ji2023meta}. Unlike federated learning (FL), which focuses on decentralized training without data exchange~\cite{rezazadeh2022specialization,zhang2022federated,erdol2022federated,zhang2023device}, meta-learning optimizes the learning process to generalize from past experiences and accelerate adaptation to new tasks. Meta-learning proves highly effective in leveraging prior knowledge to accelerate the learning process. It enables the extraction of reusable patterns and skills that support rapid adaptation to new tasks. This capability is particularly advantageous in unpredictable dynamic wireless networks, where fluctuating conditions and diverse application demands create a complex operating environment. Building on this concept, this work introduces a meta-learning-based hierarchical approach for network slicing and resource block scheduling inspired by Model-Agnostic Meta-Learning (MAML)~\cite{finn2017model}. Designed for advanced wireless O-RAN architectures~\cite{Owfi,lotfi2024meta}, this solution addresses the challenges of dynamic resource allocation and system adaptability. By minimizing performance disruptions and optimizing service delivery to user equipment (UEs), our strategy establishes a scalable framework for efficient network management in highly dynamic and demanding environments.

Building on these foundations, this paper builds upon our recent prior work~\cite{lotfi2024meta}, which focused on optimizing resource block (RB) allocation for users within the eMBB slice of an O-RAN environment. While that study primarily addressed resource management in a single slice type, this work extends the scope to a more comprehensive and realistic network slicing scenario involving multiple slice types, including eMBB, URLLC, and mMTC.
This study introduces a novel adaptive meta-hierarchical reinforcement learning (meta-HRL) framework. This framework addresses resource management across diverse slice types while incorporating efficient scheduling strategies within each slice. Meta-RL is particularly suited for rapidly changing network conditions, enabling the model to generalize across tasks and adapt quickly to dynamic environments. This capability is essential in O-RAN systems with highly variable traffic patterns and user demands. 
The proposed hierarchical approach complements this meta-RL solution by decomposing complex problems into manageable subproblems, resource allocation across slices and scheduling within slices. By structuring decisions hierarchically, the framework efficiently handles multi level resource management challenges inherent in O-RAN.
By integrating these approaches, the proposed framework ensures rapid adaptation to changing environments while maintaining scalability and decision granularity. This expanded approach significantly broadens the applicability of our earlier work, aligning it more closely with the complexities of real-world O-RAN slicing challenges. In contrast to existing learning based O-RAN solutions that often lack stability guarantees or scalability validation, our work bridges this gap through both theoretical and empirical advances. 
Specifically, we provide formal convergence and regret analysis for the proposed Meta-HRL framework, comprehensive ablation studies on the adaptive weighting mechanism, and scalability evaluations under large scale deployments. 
The main contributions of this paper are summarized as follows.
\begin{itemize}
\item We develop a novel hierarchical framework for optimizing network slicing and resource scheduling in the O-RAN system architecture. This framework integrates slicing strategies with RB allocation and downlink power optimization across multiple slice types, including eMBB, URLLC, and mMTC. The Meta-HRL approach provides a scalable solution that swiftly adapts to evolving network conditions, ensuring seamless integration of new technologies and service demands.

\item To enhance the meta-learning capabilities of our approach, this work introduces an adaptive weighting mechanism for optimizing task-specific meta-learning. This mechanism dynamically prioritizes individual learner tasks during meta-model updates, enabling better generalization across diverse tasks. This particularly benefits complex and variable environments, ensuring a more robust, flexible learning system.

\item The proposed approach emphasizes localized processing within DUs, which enhances message synchronization and mitigates delays caused by out-of-order packet arrivals. This enhancement is critical for maintaining reliability, performance, and low-latency operations in sophisticated and high-demand network scenarios, ensuring the system remains resilient to network disruptions.   

\item We address the challenges of corner cases and unpredictable demand surges in real-time scheduling. By leveraging hierarchical decision-making and meta-learning, our framework effectively mitigates performance bottlenecks in hotspots and densely populated areas, demonstrating resilience under high-load and corner-case conditions.

\item To validate the effectiveness of the proposed approach, we conduct extensive simulations within the O-RAN framework. The simulations are performed in a dynamic environment where multiple HRL agents in DUs locations manage resource allocation across heterogeneous network slices  and UEs with varying traffic patterns and QoS requirements. The results reveal a remarkable $19.8\%$ improvement in the final return value for intelligent network management compared to baseline methods. 
This demonstrates the superiority of our Meta-HRL approach in optimizing network slicing and resource allocation in dynamic and complex environments. 

\item This work establishes a new benchmark for network optimization in next-generation wireless networks, integrating adaptive decision-making, scalability, and real-time responsiveness. The proposed framework is a step forward in aligning the potential of hierarchical reinforcement learning with the demands of O-RAN architectures, paving the way for more intelligent, efficient, and resilient network systems.
\end{itemize}
\emph{To the best of our knowledge, this is the first work to propose a MAML-inspired meta-HRL framework for hierarchical network slicing and resource scheduling in O-RAN systems.}
This paper substantially extends our earlier conference version by generalizing the single slice meta learning formulation to a multi slice hierarchical reinforcement learning framework for O-RAN. 
In this extended study, the hierarchical structure explicitly separates the inter slice and intra slice decision processes, enabling scalable coordination across heterogeneous slice types. 
Furthermore, an adaptive variance weighted meta update is introduced, in which the contribution of each task to the meta-gradient is dynamically modulated through the Softmin of TD-error variance, allowing the framework to prioritize complex and high-variance slice dynamics. 
The formulation also redefines each distributed unit as an independent meta-learning task, improving generalization and adaptation under non-stationary network conditions. 
Together, these extensions enhance the scalability, robustness, and convergence stability of the original framework while offering deeper theoretical and empirical analysis. 

The paper is organized as follows: Section \ref{sec:literature} reviews related works, focusing on applying meta-learning approaches within DRL for addressing network slicing and resource allocation challenges. Section \ref{sec:system} presents the system model and formulates the problem of network slicing and resource scheduling in the O-RAN context. Our proposed Meta-HRL solution, inspired by MAML, is detailed in Section \ref{sec:MHDRL}. Section \ref{sec:analys} provides additional analysis of the proposed Meta-HRL algorithm. Simulation results and performance evaluations are discussed in Section \ref{sec:simulation}, and the paper concludes with final remarks in Section \ref{sec:conclusion}.\vspace{-0cm}
\begin{figure*}[t!]
     \centering
     \begin{subfigure}[a]{0.45\textheight}
         \centering
         \includegraphics[width=\textwidth]{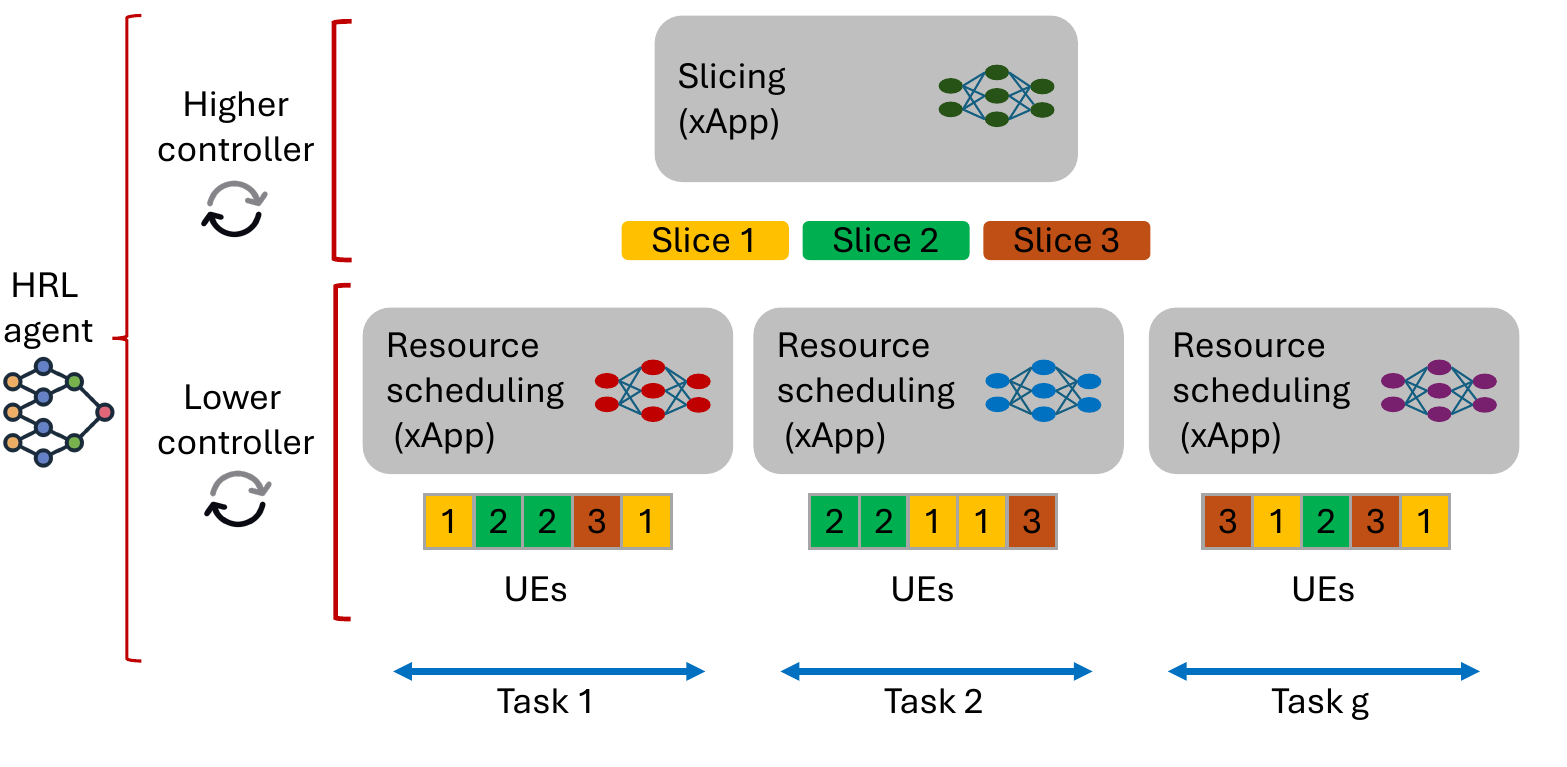}\vspace{-0cm}
    \caption{\small HRL task topology.
    }\vspace{-0cm}
    \label{HRL}
     \end{subfigure}
     \hfill
     \begin{subfigure}[a]{0.45\textheight}
         \centering
         \includegraphics[width=\textwidth]{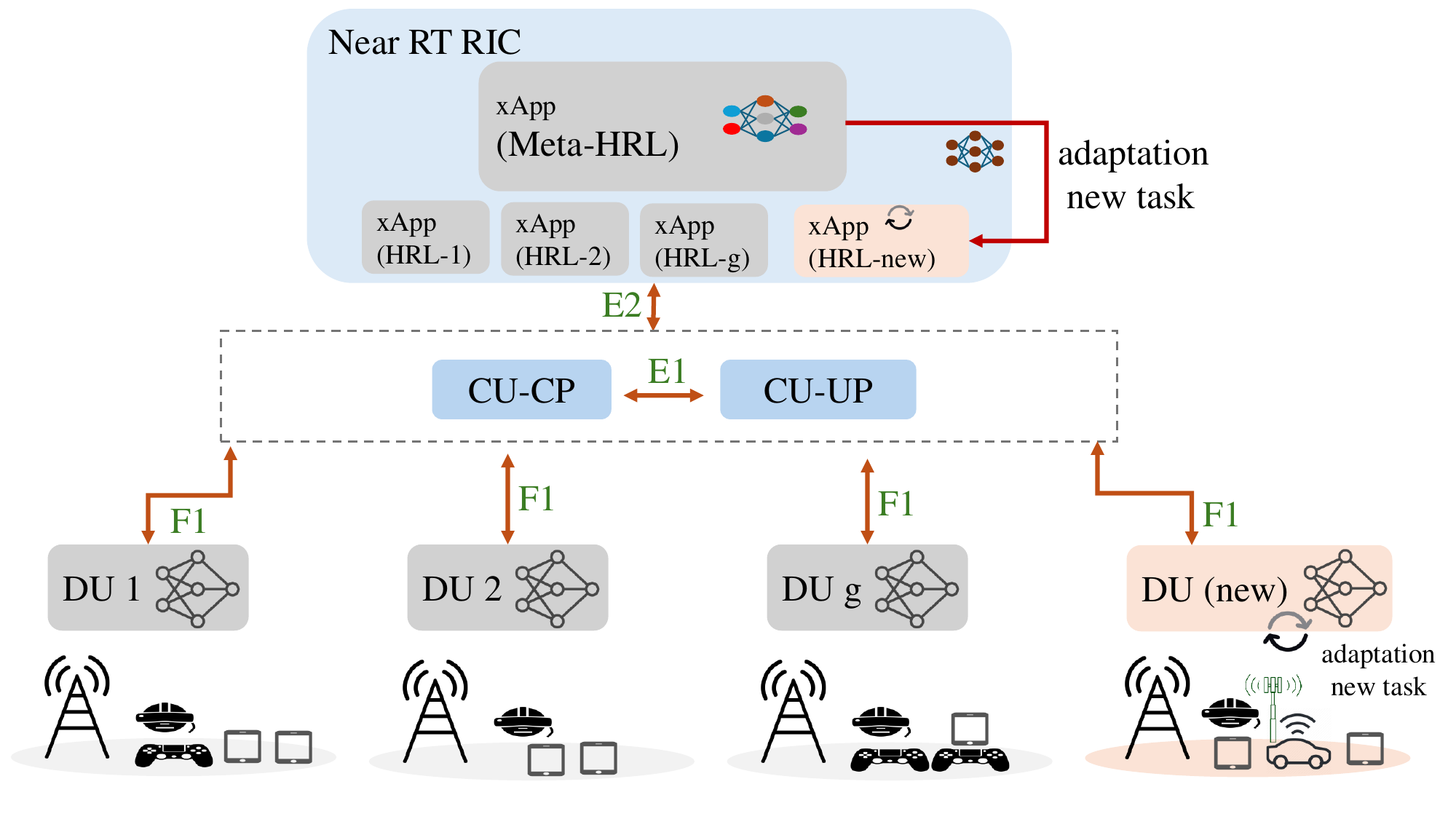}\vspace{-0cm}
    \caption{\small MAML-HRL network framework in adaptation phase.
    }\vspace{-0cm}
    \label{MAML-HRL-net}
     \end{subfigure}
        \caption{System model topology for the proposed MAML-HRL resource management in O-RAN architecture.
       }
        \label{sys_graph}
        \vspace{-0.cm}
\end{figure*}

\section{Related Works}\label{sec:literature}
Recent advancements in wireless networks have highlighted the need for innovation in O-RAN architecture, with contemporary research increasingly focusing on the integration of intelligent frameworks such as meta-learning and DRL to enhance network efficiency and adaptability.

\subsection{Resource Allocation and Slicing Frameworks} 
The work in~\cite{chen2023hierarchical} proposes a meta-MDP framework to address resource allocation challenges in O-RAN by employing a two-timescale HRL approach. This framework decouples resource management into a long-timescale slice-level orchestration problem, optimized with deep RL, and a short-timescale user-level configuration problem solved using a linear-decomposition-based meta-RL algorithm. While their approach effectively balances slice-level and user-level decisions, the reliance on a fixed two-timescale structure may limit adaptability in highly dynamic or unpredictable network conditions. Furthermore, their focus on linear decomposition for user-level configurations could reduce flexibility when dealing with non-linear interactions in complex environments.

Similarly, \cite{rezazadeh2022specialization} proposes a federated DRL (FDRL) framework for dynamic slicing in 6G networks. This approach distributes learning tasks to local agents (e.g., for edge or cloud slices) while maintaining a global policy for coordinated decision-making. The paper emphasizes scalability and adaptive learning in decentralized environments, allowing agents to specialize in their slice-specific tasks and reducing communication overhead. In another study,~\cite{li2018deep} applies DRL techniques to manage resource allocation in network slicing. By using deep Q-learning, this approach optimizes radio and core resources, improving spectral efficiency and service quality. It dynamically allocates resources to different slices, prioritizing traffic demands based on QoE metrics and network conditions. 
Moreover, \cite{zhang2022federated} utilizes federated DRL to coordinate multiple xAPPs, focusing on power control and slice-based resource allocation in O-RAN. However, their approach assumes fixed action and state spaces, which may not generalize well to diverse and dynamic O-RAN scenarios. Additionally, work in~\cite{zhou2022learning} introduces deep transfer RL for joint resource allocation, leveraging knowledge from expert agents to improve learning efficiency. 

\subsection{DRL and Multi-agent Strategies} 

The studies in \cite{lotfi2024open,kavehmadavani2024empowering} introduce a multi-agent DRL approach integrated with Long short-term memory (LSTM) for prediction and optimization, focusing on network resource management and traffic steering in Open RAN. Specifically, \cite{kavehmadavani2024empowering} emphasizes balancing traffic demands and improving system performance in multi-traffic scenarios, While \cite{lotfiattention, lotfi2024open} explores a distributed DRL approach within the O-RAN system architecture, it optimizes xApp resource allocation in the near-real-time (near-RT) RIC, incorporates an LSTM-based prediction module as an rApp in the non-real-time (non-RT) RIC~\cite{lotfi2024open}, and employs attention value weights in the global critic network of the xApp to prioritize the values of distributed DRL agents~\cite{lotfiattention}. 
Additionally, work in \cite{habib2023hierarchical} leverages hierarchical reinforcement learning (HRL) for traffic steering in multi-radio access technology (RAT) environments. By separating control into high-level meta-controllers for overarching policies and low-level controllers for detailed actions, this hierarchical structure enhances decision-making efficiency and system performance, particularly in large-scale deployments. Similarly, \cite{lee2021ran} implements an RL model for personalized network optimization, focusing on retraining with local data for cell-specific improvements.

Moreover, \cite{lotfi2022evolutionary} presents a hybrid framework combining evolutionary algorithms with DRL to optimize resource allocation across slices, including eMBB, URLLC, and mMTC. By leveraging the exploration capabilities of population-based evolutionary algorithms and the sample efficiency of DRL, the framework achieves faster convergence and robust policy learning in dynamic and resource-constrained environments. Furthermore, \cite{erdol2022federated} proposes a federated RL-based traffic steering algorithm within the non-RT RIC for RAT allocation, using Deep Q-Networks (DQN) to dynamically steer traffic across RATs, thereby enhancing network performance and user experience. While federated RL supports distributed learning, meta-RL offers superior adaptability and scalability in dynamic O-RAN scenarios, making it a more suitable solution for seamless implementation. Finally, our work in \cite{lotfi2024meta} introduces a Meta-DRL framework inspired by MAML to optimize resource block and downlink power allocation in O-RAN systems. The approach leverages O-RAN's disaggregated architecture by deploying distributed agents at DUs for localized decision-making, enabling real-time adaptability to dynamic network conditions. While this study effectively addresses resource allocation for eMBB slices, it focuses on single-slice optimization.

\subsection{Advanced Learning Techniques for Optimization} %
Work in \cite{nagib2023accelerating} investigates transfer learning to accelerate RL convergence in dynamic RAN slicing by introducing a predictive approach for policy reuse, addressing slow convergence challenges in live networks. Similarly, \cite{niu2023multiagent} presents a multi-agent meta-RL framework using Proximal Policy Optimization (PPO) for task scheduling in edge computing environments. While the meta-learning component enables agents to adapt to varying task types, the proposed approach is confined to static edge computing setups and does not address the complexities of dynamic, multi-slice network slicing scenarios in O-RAN systems. Furthermore, it lacks a hierarchical decision-making structure, which is essential for managing both slice-level orchestration and user-level resource allocation in complex network environments. 
In addition, \cite{zhang2023device} proposes a federated learning framework for UE-centric traffic steering, where UEs utilize local observations and a global model for decision-making. The framework leverages attention-weighted knowledge transfer and model compression to enhance scalability and performance. Similarly, \cite{zhou2022knowledge} explores AI-enabled knowledge transfer and reuse to improve RAN slicing, with a focus on reducing training efforts through transfer learning.


While the application of meta-MDP in network slicing, as demonstrated in recent studies, provides effective handling of specific hierarchical scenarios, it lacks the broad adaptability inherent to meta-RL strategies. Unlike meta-RL, which excels in applying learned strategies across diverse tasks due to its task-agnostic nature, meta-MDP focuses on optimizing a single, structured task. This specialization can lead to efficient solutions within narrowly defined scenarios but often comes at the cost of reduced flexibility and scalability. Consequently, meta-MDP frameworks are limited in dynamic environments where conditions and requirements frequently change, such as in O-RAN systems with varying traffic demands and service types. These distinctions highlight the fundamental trade-off between specialized precision and generalized adaptability in network slicing methodologies. To address these limitations, in this study, we propose a novel adaptive meta-HRL framework that delivers the adaptability and scalability required for hierarchical network slicing and resource scheduling in dynamic and unpredictable O-RAN environments.
\section{SYSTEM MODEL AND PROBLEM FORMULATION}\label{sec:system}

\subsection{Overall Architecture}
In a O-RAN wireless network, we have $L=3$ multiple network slices designed to serve different types of services, such as enhanced Mobile Broadband (eMBB) and Ultra-Reliable Low-Latency Communications (URLLC). Each slice has unique Quality of Service (QoS) requirements. The network dynamically adjusts by potentially adding or turning on/off DUs based on user demand. In addition, the network can also incorporate DU-RU (Radio Units) pairs, such as small cells. These small cells are particularly useful for addressing localized demand spikes and can be dynamically activated or deactivated to optimize resource utilization. For example, a new DU could connect to a small cell instead of a macro-cell, offering a scalable and efficient solution for localized coverage and capacity enhancements. 
This dynamic nature involves two key resource management steps: slicing to allocate network resources to different services and scheduling to distribute these resources among individual users within a slice. In this architecture, we consider several Central Units (CUs), DUs, and RUs. These elements are coordinated by RIC to dynamically manage system resources. Here, we have an inter-slice manager, which manages resources across different network slices, and intra-slice manager which manage resources within a single slice. Figure \ref{sys_graph} shows our proposed scenario.  
\subsection{Achievable Data Rate}
The O-RAN architecture illustrated in Figure \ref{sys_graph} employs various network slices, each with different QoS criteria, to serve diverse services of UEs. The near-RT RIC module oversees all slices and resources, utilizing dynamic resource management to adjust to fluctuating wireless channels. However, the stochastic nature of this dynamic approach complicates resource allocation. An orthogonal frequency-division multiplexing (OFDM) transmission scheme is employed to avoid intra-cell interference~\cite{3gpp15}. The achievable data rate for each UE $u$, associated with slice $l$ and assigned RU-DU $\rho$, considering a Rayleigh fading transmission channel and assuming the system operates within discrete time frames $t$, can be expressed as:
\begin{align}\label{urate} 
    c^l_{u,\rho}(t) = \sum_{k=1}^{K^l_{\rho}} B e_{u,k} b_{l,k} \log \Big(1+\frac{p_{u} d_{u,\rho}(t)^{-\eta} |h_{u,k}(t)|^2}{ I_{u,k}(t)+ \sigma^2}\Big),
\end{align}
where $e_{u,k} \in \{0, 1\}$ is a binary variable indicating RB allocation for user $u$ in RB $k$,  $b_{l,k}$ is RB distribution indicator for slice $l$ in RB $k$, and $B$ represents reach RB bandwidth. 
Let $p_{u}$ denote the transmission power allocated to resource block $k$ of RU $\rho$. The variable $d_{u,\rho}(t)$ represents the distance between RU $\rho$ and UE $u$ at each time frame, which varies over time due to the mobility of the UEs. Additionally, $\eta$ represents the path loss exponent, and $|h_{u,k}(t)|^2$ indicates the time-varying channel gain due to Rayleigh fading for each time frame. In equation \eqref{urate}, the term $I_{u,k}(t)=\sum_{\rho' \neq \rho} \sum_{u'\neq u} e_{u',k} p_{u'} d_{u',\rho'}(t)^{-\eta} |h_{u',k}(t)|^2$, indicates the inter-cell interference of downlink transmission from other RUs on RB $k$, and  $\sigma^2$ is the additive white Gaussian noise (AWGN) variance.

\subsection{Problem Formulation}
The primary objective is to maximize the overall network throughput while ensuring the specific QoS requirements of each slice are met. This is represented as:
\begin{subequations} 
\begin{align}\label{opt1}
 \argmax_{\boldsymbol{b},\boldsymbol{e}} & \hspace{0.5cm} 
  \sum_{l}\sum_{\rho} c^l_{u,\rho}(\boldsymbol{b},\boldsymbol{e}),\\
 \text{s.t.,} 
 & \hspace{0.5cm}  \boldsymbol{e} \in \{0,1\}^{N_u \times K^l_\rho}, \\
& \hspace{0.5cm}  \sum_{u=1}^{N_\rho}\sum_{k=1}^{K_\rho}e_{u,k} \leq K_\rho, \label{opt1_rb}\\
& \hspace{0.5cm} C^l_{u}\geq C^l_{\text{min}} .\label{opt1_qos} 
\end{align}
\end{subequations}
In this formulation, several constraints are defined to manage network resources effectively. These include constraints on RB availability as outlined in~\eqref{opt1_rb}, 
and distinct throughput constraints in each slice, described in~\eqref{opt1_qos}. These constraints ensure that specific throughput demands are met in each slice. $C^l_{u}$ represents each user throughput in slice $l$ and quantifies how well the current resource allocation meets the demands of slice $l$. 
We consider it as a constraint to ensure that the performance of each slice does not fall below the minimum acceptable level, $C^l_{\text{min}}$, thereby maintaining adequate service quality for all users and applications associated with that slice. 
However, the complex and dynamic nature of modern wireless networks requires an advanced approach to resource management that can adapt to varying conditions and user demands. Our formulation of the resource allocation challenge as a mixed-integer nonlinear programming problem seeks to optimize resource distribution both globally, across different network slices, and locally, within each slice. Given this complexity, traditional optimization techniques alone may fall short in effectively addressing the rapidly changing network conditions. This is where the integration of meta-learning and hierarchical reinforcement learning (HRL) becomes invaluable.

\section{Proposed Hierarchical Meta-RL Resource Distribution}\label{sec:MHDRL}
\subsection{Framework Overview} 
Meta-learning, or learning to learn, enhances the system’s ability to quickly adapt to new scenarios based on past experiences, ensuring that the network can respond efficiently to sudden changes in demand, such as a spike in user activity or the need to activate additional DUs\cite{finn2017model}. By leveraging past experiences, the system can fine-tune its resource allocation strategies, maintaining optimal performance and ensuring that the QoS-related performance metric $C^l_{\boldsymbol{p},\boldsymbol{e}}$ remains above the minimum acceptable level $C^l_{min}$ for each slice. 
Moreover, HRL provides a structured approach to manage resource allocation by decomposing the problem into a hierarchy of simpler sub-problems. The higher controller level involves slicing the network to allocate resources to different services, while the lower controller level focuses on scheduling resources within each slice to individual users. By combining HRL with meta-learning, our proposed solution can dynamically adjust to varying network conditions, ensuring that resources are allocated optimally both globally and locally. This synergy between meta-learning and HRL not only enhances the adaptability and efficiency of the network but also provides a robust framework for handling the complexity and variability inherent in modern wireless communication systems. 

\begin{figure}[t!]
         \centering
         \includegraphics[width=0.85\columnwidth]{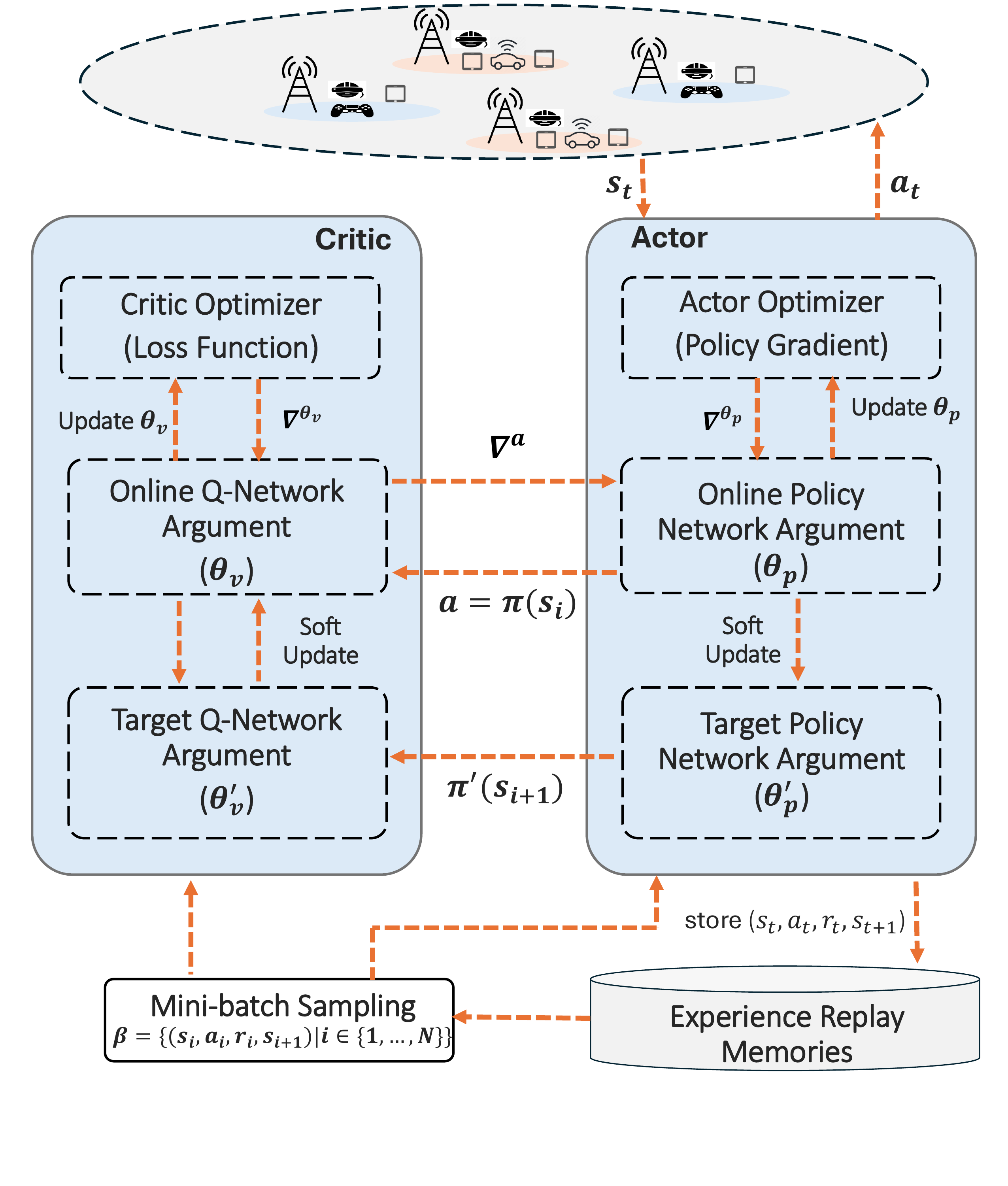}\vspace{-0cm}
    \caption{\small Flowchart of the DDPG algorithm.
    }\vspace{-0cm}
    \label{ddpg_flow}
     \end{figure}

Building on this synergy between meta-learning and HRL, our approach introduces a hierarchical framework to intelligently and dynamically manage resource blocks among UEs assigned to different network slices. At the higher controller level, implemented as an xApp in the non-RT RIC, a crucial decision-making process determines how to distribute shared resources among slices, leveraging the global network state and performance to learn an optimal resource distribution policy. This stage significantly influences local allocation decisions at the lower controller level, also deployed as an xApp, where resources are adaptively distributed among UEs within each slice based on their specific QoS demands. Both levels employ the Deep Deterministic Policy Gradient (DDPG) technique to optimize their respective tasks. 
DDPG is an actor-critic model-free and off-policy algorithm. The actor and critic networks iteratively learn the policy and a value networks as shows in Fig.~\ref{ddpg_flow}. 
The critic network parameterized by $\boldsymbol{\theta}_v$ will be updated by minimizing the temporal difference (TD) error using the Bellman equation: 
\vspace{-0.cm}
\begin{align}\label{critic_update}
    \min_{\theta_v} \frac{1}{e}\sum_{i=1}^{e} \bigg(r_i + \gamma Q_{\pi_{i+1}}(s_{i+1},a_{i+1};\theta_v)-Q_{\pi_i}(s_{i},a_{i};\theta_v)\bigg)^2. \vspace{-0.cm}
\end{align}
where $Q_{\pi_{i+1}}$ is the target Q-value, and $\gamma$ is the discount factor.
The actor network parameterized by $\boldsymbol{\theta}_p$ is updated to maximize the Q-value by following the policy gradient with $e$ random samples transitions: 
\begin{align}\label{actor_update}
    \nabla_{\theta_p}J = \frac{1}{e}\sum_{i=1}^{e} \nabla Q\big(s_i,a_i;\theta_v\big) \nabla_{\theta_p}\pi\big(a_i,s_i;\theta_p\big).
\end{align}
where $\nabla_{\theta_p}\pi$ represents the gradient of the policy network $\pi$ with respect to its parameters $\theta_p$. DDPG effectively learns policies for tasks involving continuous actions by iteratively updating the actor and critic networks. In our framework, these updates are applied independently at higher and lower levels to optimize global slicing and local scheduling tasks.
To further enhance the adaptability and efficiency of this hierarchical framework, we integrate meta-learning over this HRL structure. This combination is crucial because it enables the system to rapidly learn and adapt to new tasks or agents in response to changing network's UEs demands and evolving condition.
Each task in the meta-Reinforcement Learning (meta-RL) approach is a multi-level RL scenario encompassing both inter-slice and intra-slice resource management. Different actors are located in each DU across the network, experiencing various UE traffics and service demands. For example, one DU might experience a higher demand for eMBB slice due to a large concert or sporting event, necessitating increased bandwidth and low-latency for live streaming services. Meanwhile, another DU might see a surge in UE connections during a tech conference, requiring reliable and high-capacity connections for numerous device interactions and data exchanges. Additionally, one DU might face higher UE dynamicity in a transportation hub like a busy airport, where users are constantly moving and connecting to the network, whereas another DU might serve a residential area with more stable and predictable usage patterns, such as consistent streaming and web browsing demands. Therefore, their resource management policies must adapt to these diverse and dynamic conditions. By integrating meta-learning with HRL, our approach allows these DUs to develop and refine resource allocation strategies that are tailored to their specific circumstances. This ensures that each DU can effectively manage its resources based on the unique demands of its assigned UEs, leading to improved overall network performance and responsiveness. Our solution provides a robust framework for modern wireless networks, enabling efficient and adaptive resource management in the face of varying and evolving demands.

Each slice QoS is considered as specific comprehensive KPI metrics as $\mathcal{M} =\{\mu_r,\textit{d}_s,\textit{l}_d\}$, where they represent network aggregated throughput, network UE density support, and network maximum latency, respectively. 
For slice 1 (eMBB), the KPI focuses on aggregated user throughput to meet the high data rate demands of Enhanced Mobile Broadband (eMBB) services. Defined as 
\begin{align}\label{Q_embb}
    \mu_r = \frac{1}{N_u}\sum_{i=1}^{N_u} C_i, 
    \end{align}
where $C_i$ is the throughput rate for UE $i$, with $N_u$ as the total number of UEs. This metric supports bandwidth-intensive applications like high-definition video streaming and virtual reality. 
For slice 2 (mMTC), the KPI is defined as a Quality-Weighted Network Capacity, which integrates user throughput with a coefficient reflecting UE density support to balance device connectivity and service quality. The KPI is formulated as 
\begin{align}\label{Q_mmtc}
     \textit{d}_s = \frac{\sum_{i=1}^{N_u} \mathds{1}_{(C_i > \lambda_i)}}{N_u} \sum_{i=1}^{N_u} C_i,
\end{align}
where $\lambda_i$ represents the threshold value for the acceptable throughput rate of the $i$-th UE, and $\mathds{1}$ is the indicator function that equals $1$ if $C_i(t) \geq \lambda_i $, and $0$ otherwise.  
This metric is crucial for IoT applications and dense urban environments, where supporting a high number of devices is essential to ensure efficient connectivity and reliable service delivery. 
For slice 3 (URLLC), the KPI prioritizes latency, measured as the maximum user transmission delay, represented by 
\begin{align}\label{Q-URLLC}
    \textit{l}_d = \max (\tau_i), \forall i \in N_u,
\end{align}
where $\tau_i$ is the latency experienced by UE $i$. This metric is essential for Ultra-Reliable Low-Latency Communications (URLLC), where minimal latency is critical for applications such as autonomous driving and industrial automation, ensuring timely and reliable data transmission. Therefore, $Q_l$ in each slice is $\mathcal{M}[l]$, for $ l = \{1, 2, 3\}$.
\subsection{Hierarchical-RL Network Slicing and Resource Scheduling}
Traditional reinforcement learning (RL) frameworks approach resource allocation as a single-agent problem, where an agent interacts with the environment to maximize long-term rewards, defined by a Markov Decision Process (MDP) $\langle \mathcal{S}, \mathcal{A}, T, R \rangle$. While effective for simpler scenarios, this approach struggles with the complexity of modern network slicing and resource scheduling tasks, which require balancing diverse and dynamic QoS requirements across slices.
Hierarchical reinforcement learning (HRL) addresses these challenges by introducing a hierarchical structure to the RL paradigm, enhancing scalability and efficiency. In HRL, the MDP is extended to include a goal space $\mathcal{G}$, transforming it into $\langle \mathcal{S}, \mathcal{A}, T, R, \mathcal{G} \rangle$. The framework comprises two levels of decision-making: a higher controller, which determines high-level goals or sub-goals for optimization, and a lower controller, which executes fine-grained resource scheduling based on these goals. This hierarchical structure allows the higher controller to focus on global objectives, such as slice-level QoS satisfaction, while the lower controller handles local decisions, such as specific RB allocations.
By leveraging this multi-level architecture, the proposed HRL framework significantly enhances learning efficiency and adaptability in O-RAN. The hierarchical decomposition enables better management of diverse functionalities, improving resource utilization and ensuring compliance with global QoS metrics. 
The higher controller agent’s state space $\mathcal{S}^h$, action space $\mathcal{A}^h$, and reward function $R^h$ are defined as follows.

\subsubsection{Higher-controller MDP} The higher-controller as global controller is responsible for the high-level policies for the agent. Its state space $\mathcal{S}^h$, action space $\mathcal{A}^h$, and reward function $R^h$ of the global-controller are defined as follows.

\begin{itemize}
\item{Higher state:}
At each time step, the current status of the network is captured by $s_t \in \mathcal{S}^h$. This includes details such as the QoS achieved by each slice, $Q_l$, the number of UEs per slice, $N_l(t)$, and the last resource allocation action, $a_{t-1}$. Therefore, the observation available to the intelligent agent at time $t$ can be detailed as $s_{h,t} = \big\{Q_l, N_l, a_{t-1} | \forall l \in \mathcal{L}\big\}.$

\item{Higher action / Goal for lower-controller:}
At each time step $t$, the vector $a_t \in \mathcal{A}^h$ indicates the resources needed for the O-RAN slices. Consequently, the agent employs its policy to decide on the right course of action, defining $a_{h,t} = \boldsymbol{b}$.

\item{Higher reward: }
 The DRL agent learns to maximize a global reward function $R_g$, which is composite measure of network QoS across all slices according to following:
\begin{align}\label{reward_global}
    R_h(t)= \sum_{l} Q_l (t).
\end{align}
\end{itemize}
\subsubsection{Lower-controller MDP} The lower controller is responsible for executing lower-level policies to optimize resource scheduling for individual slices. At each timestep, it receives instant goals from the higher-controller and focuses on allocating resources among users within the slice. Its MDP components are defined as follows:

\begin{itemize}
\item{Lower state:}
At timestep $t$, the state $s_t \in \mathcal{S}^l$ reflects the O-RAN status, including users' QoS metrics as average ($Q_a$), minimum ($Q_m$), and maximum ($Q_x$) along with the most recent resource allocation $\boldsymbol{a}_{t-1}$. Here, QoS is measured as users' throughput. 
Thus, the agent's observation is given by $s_t = \{Q_a, Q_m, Q_x, \boldsymbol{a}_{t-1} \mid \forall u \in N_\rho\}$. 

\item{Lower action:} 
At each timestep $t$, the action vector $a_{l,t} \in \mathcal{A}^l$ specifies the number of resource blocks assigned to each user. The agent's policy determines the optimal action, represented as $a_{l,t} = {\boldsymbol{e}}$, where $\boldsymbol{e}$ is an indicator vector for RB assignments. 

\item{Lower reward:} 
The reward function balances resource efficiency and QoS satisfaction. It rewards policies that optimize resource utilization while penalizing excessive RB usage. The penalty for over-allocation is defined as $\Tilde{K}_r = \max(0,\sum_{u=1}^{N_\rho} \sum_{k=1}^{K_\rho} e_{u,k}-K_\rho) $. The total reward $r_t$ combines a sigmoid function of the normalized minimum QoS $\Tilde{Q}_m$, 
and the penalty $\Tilde{K}_r$ for excessive RB consumption: 
\vspace{-0.cm}
\begin{equation}
    r_t = \text{sigmoid}(\Tilde{Q}_m) - \text{sigmoid}(\Tilde{K}_r), 
\end{equation}
where $\Tilde{Q}_m = (Q_m - c_m)/(c_x - c_m)$ normalizes $Q_m$ 
based on minimum and maximum values of service demands as $c_m$, $c_x$, respectively. 
\end{itemize}
This hierarchical approach lays the groundwork for intelligent resource allocation, however, the dynamic and diverse nature of network conditions necessitates a system capable of quickly adapting to changing demands. To address this, we extend our framework with an adaptive meta-learning-based enhancement, detailed in the next section.
\begin{figure}[t!]
         \centering
         \includegraphics[width=0.88\columnwidth]{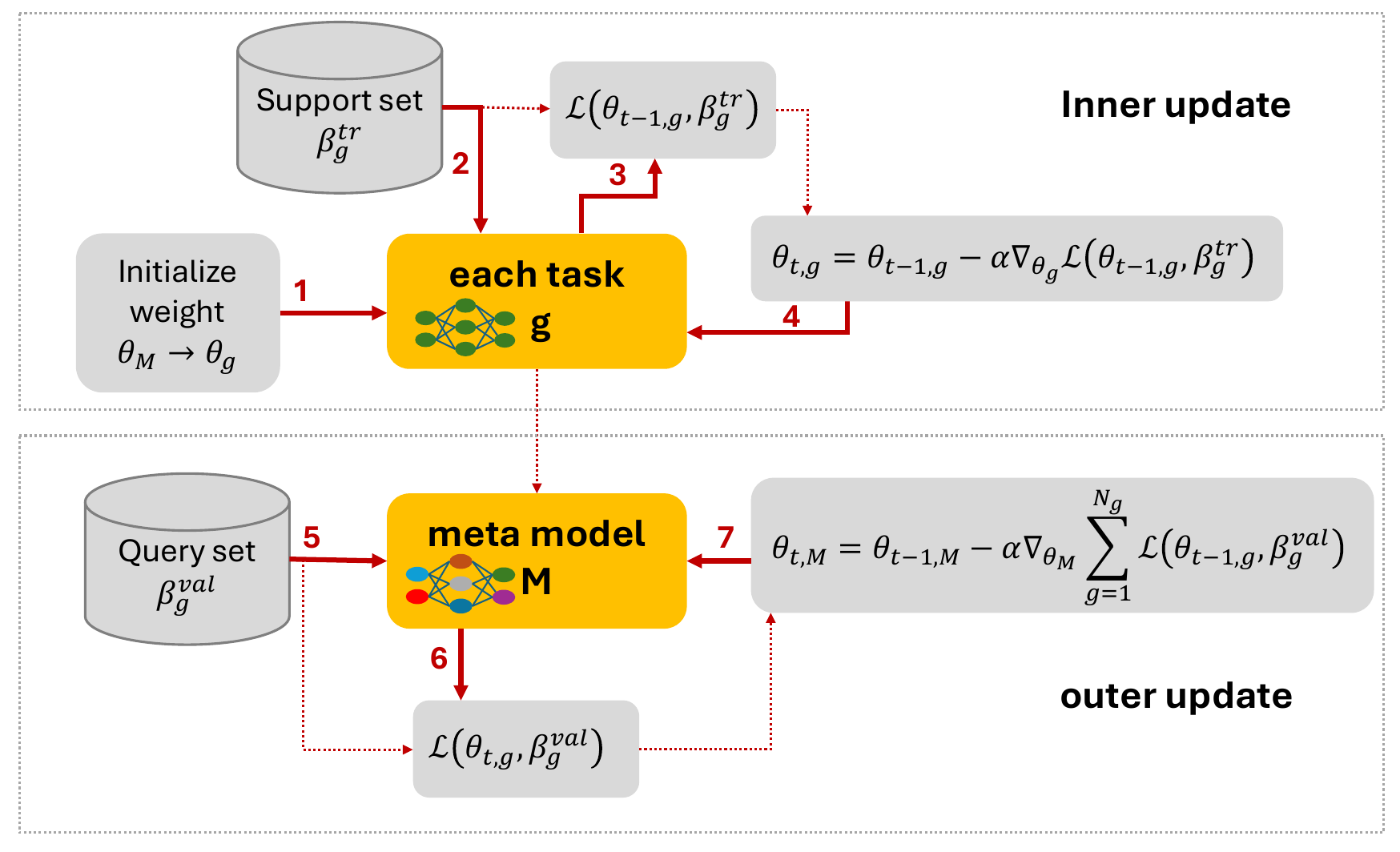}\vspace{-0cm}
    \caption{\small MAML-HRL training framework.
    }\vspace{-0cm}
    \label{maml-training}
     \end{figure}
\subsection{Meta learning-based HRL} 
To further enhance the adaptability of the hierarchical framework, we integrate the MAML approach with HRL. For each environment condition, a meta-HRL model rapidly adapts to the unique resource allocation requirements, considering the task-specific constraints and QoS goals. This ability to quickly fine-tune policies is crucial for responding to real-time variations in network conditions and dynamic user demands, ensuring efficient and responsive resource management. 
The proposed meta-learning approach aims to train $N_g$ distinct HRL agents as separate tasks, each deployed in a distributed DU and implemented as an xApp running in the non-RT RIC. These local HRL tasks are optimized to manage slicing and RB scheduling within their respective DUs. At the global level, a centralized meta-HRL agent xApp acts as a meta-model, coordinating the collaboration between these distributed xApps to ensure better generalization across diverse network scenarios. 
This hierarchical design leverages HRL for managing complex, coupled tasks locally while enabling the meta-HRL agent to provide global coordination and scalability. By centralizing the meta-HRL  process, the system ensures efficient policy transfer, rapid adaptation to new DUs condition, and a unified optimization framework to handle task coupling, reduce conflicts, and effectively scale. This design is illustrated in Fig. \ref{MAML-HRL-net}.

The meta-HRL agent is designed to adapt to new tasks or environments with minimal training rapidly. Each HRL agent operates in distinct environments and has its own state space $\mathcal{S}$, action space $\mathcal{A}$, and customized reward function $R_g$ to ensure they are well-suited to address diverse conditions and dynamic demands across network locations. When designing meta tasks, the goal is to prepare the model for real-world challenges that vary a lot from one situation to another. For example, a meta task might involve handling sudden changes in network demands or learning to adapt when resources are unexpectedly limited. These scenarios help train the model to respond effectively without needing a lot of extra training~\cite{finn2017model}. The task-specific reward function is defined as 
\begin{align}
    R_g = \{r_t | \Tilde{Q}_m = \frac{Q_m - c_{m,g}}{c_{x,g} - c_{m,g}}\}, \forall g \in [0, N_g],
\end{align}
where $c_{m,g}$ and $c_{x,g}$ represent unique parameters tailored to each environment. In this context, the HRL agent $g$ utilizes a support set $\mathcal{B}^{tr}_g$, which consists of data samples specific to each task. This set trains the model within the task and adapts it to the given environment. A query set $\mathcal{B}^{val}_g$ is employed to evaluate the model's performance and update the meta-HRL model's parameters during training. The query set serves as a validation set to compute the meta-objective, which represents the loss based on the predictions made on the query set. The following equations outline meta HRL tasks and meta-HRL model network parameters updating as $\theta_{t,g}$ and $\theta_{t,M}$, respectively, and ensuring that the meta-HRL agent improves its generalization capability across diverse tasks.
\begin{align}
    \theta_{t,g} = \theta_{t-1,g} - \alpha \nabla_{\theta_g}\mathcal{L}(\theta_{t-1,g}, \beta^{tr}_g), \forall g \in [0,N_g], \label{update_agent_t}\\
    \theta_{t,M} = \theta_{t-1,M} - \alpha \nabla_{\theta_M}\sum_{g=1}^{N_g}  \mathcal{L}(\theta_{t-1,g}, \beta^{val}_g), \label{update_meta_t} 
\end{align}
where $\mathcal{L}(\theta_{t,g})$ denotes the loss function of the DDPG algorithm for agent $g$ at time step $t$, where $\theta_{t,g}$ represents the policy or value network parameters of agent $g$, and $\theta_{t,M}$ represents the policy or value network parameters of the meta-HRL model $M$ at time step $t$. Accordingly, equation \eqref{update_agent_t} is responsible for updating the policy or value parameters of the individual HRL agent using the gradient descent approach with $\alpha$ learning rate. Meanwhile, equation \eqref{update_meta_t} updates the meta-model policy or value parameters by integrating the adaptation capabilities demonstrated by the trained agents across different tasks. Fig.~\ref{maml-training} represents this process. 

\subsection{Training Phase}
The training phases for integrating and coordinating the HRL slicing and scheduling agent with the meta-HRL approach for resource management are outlined in the following steps. 
\subsubsection{Initial Independent Training of the HRL Slicing and Scheduling Agent}
In this phase, the focus is on training the HRL agent for slicing and scheduling networks to make competent and stable decisions on how network resources are divided among slices and UEs. This training is conducted independently to ensure that the resource allocation decisions are robust and effective, providing a solid foundation for the subsequent steps. After this training step, we ensure that the HRL agent can develop a reliable set of policies for resource partitioning without interference from the complexities of resource allocation.
\subsubsection{Integration and Training of the Meta-HRL agent}
Once the HRL agent for slicing and scheduling is trained to have a stable resource allocation policy, the next step of training starts. This phase starts with the meta-HRL agent learning to allocate resources effectively within the slices and UEs based on demand. The purpose here is to allow meta-HRL agent to develop a solid understanding of various slicing and scheduling and resource allocation challenges, using the stable input provided by the resource allocation decisions. This helps in building a strong base of knowledge before tackling more dynamic or unpredictable scenarios. 
\subsubsection{Dynamic Introduction of New Tasks and Online Learning}

In this phase, each new HRL agent introduced is treated as a distinct task within the meta-HRL framework, requiring customized resource allocation and scheduling strategies. When a new HRL agent is created, it initiates a specific task for the meta-HRL agent, starting with an initial set of parameters reflecting it's unique demands. The meta-HRL agent then engage in real-time learning, using initial operational data from the new environment to adapt and refine resource allocation strategies. As more data become available, these strategies are continuously updated, ensuring optimal resource management tailored to the evolving conditions of each new HRL agent task. This approach allows the meta-HRL framework to dynamically adjust to new scenarios, maintaining efficient network operations and high service quality.

\subsubsection{Continuous Adaptation}

In the final phase, the entire system, both the HRL agents and the meta-HRL agent, are subject to ongoing monitoring and evaluation. This continuous oversight helps identify any adjustments that might be needed, whether in slicing parts or resource scheduling training. This final step by performing fine tuning, ensure that both the slicing strategies and resource allocation policies remain aligned with the network’s objectives and performance standards.

The stability of the two-level learning process and its convergence properties are analyzed in Section~\ref{sec:theory}, followed by large-scale and ablation evaluations in Section~\ref{sec:scalability} and Section~\ref{sec:ablation}.

\section{Adaptive Meta-HRL}
\subsection{Overall Framework}
As discussed our proposed meta-HRL approach in the previous section, this section presents an additional improvement to further enhance the approach performance and efficiency. In traditional MAML-inspired meta-HRL frameworks, each HRL agent contributes equally to the training of the meta-HRL model. However, by adopting an adaptive meta-HRL approach, we can dynamically adjust our learning strategies based on real-time feedback and evolving environmental conditions. To optimize this process, dynamic weights are incorporated into the meta-HRL model's loss function update, enabling the model to prioritize tasks with higher complexity and variability. These weights ($\omega_g$) are tailored based on task-specific metrics such as task variance, which quantifies the unpredictability or range of outcomes within a task. This adaptability enhances the model's responsiveness and performance in addressing complex, non-stationary problems. 

Task variance, which measures the variability of outcomes or observations within a task, plays a critical role in assigning dynamic weights. High task variance indicates tasks with unpredictable dynamics or diverse potential outcomes, making them key targets for adjustment. By incorporating this metric, the learning process becomes more focused and the meta-model better adapts to complex and dynamic environments. The updated meta-model parameters in \eqref{update_meta_t} are transformed into an adaptive version as shown in the following equation.
\begin{align} 
\theta_{t,M} &= \theta_{t-1,M} - \alpha \nabla_{\theta_M}\sum_{g=1}^{N_g} \omega_g \cdot \mathcal{L}(\theta_{t-1,g}, \beta^{val}_g). \label{update_meta_adaptive} 
\end{align}
This dynamic adaptation enables the framework to address complex, non-stationary problems and evolving environmental conditions effectively. By integrating task variance into the meta-objective, the model tailors its learning strategies to the complexity and real-time demands of each task, thereby enhancing overall performance and ensuring efficient resource management. Furthermore, by dynamically adjusting only to tasks with significant variability, the model avoids over-adaptation in stable conditions, preventing catastrophic forgetting and reducing unnecessary computations.
\subsection{Adaptive Weight}
Metrics that measure the divergence between expected and actual outcomes are valuable for reflecting task complexity. The temporal difference error or TD-error is particularly useful, as it measures the difference between the estimated values of states and their actual outcomes, defined as
\begin{align}\label{td_err}
    \delta_{TD} = R_{t+1} + \gamma V(s_{t+1}) - V(s_t), 
\end{align}
where $R_{t+1}$ is the reward received after moving to state $s_{t+1}$, $\gamma$ is the discount factor, and $V(s)$ is the estimated value of state $s$. Variance in TD-error, $\sigma^2_{TD}$, is particularly telling, indicating tasks that may be more complex or stochastic. This variance not only serves to reflect the complexity and unpredictability of tasks that a higher variance suggests more unpredictable rewards and state transitions, but it also allows for adaptive weighting in the training process. 
We define adaptive weights for each agent as 
\begin{align}\label{w_adapt}
\omega_g = \text{Softmin}(\sigma^2_{TD_g}),
\end{align}
where $\text{Softmin}$ is a function that emphasizes smaller values, allowing them to dominate the output probabilities. The $\text{Softmin}$ function is defined as follows:
\begin{align}\label{softmin}
    \text{Softmin}(\sigma^2_{TD_g}) = \frac{e^{-\sigma^2_{TD_g}}}{\sum_j e^{-\sigma^2_{TD_j}}},
\end{align}
where $\sigma^2_{TD_g}$ represents the variance of the TD error for the $g$-th HRL agent. 
We can dynamically adjust the focus on each HRL agent's performance by integrating TD error variance into the importance vector. Agents that are exhibiting higher variance require more attention to ensure effective convergence and the learning of stable policies. 
Furthermore, in a meta-learning context, leveraging TD error variance helps prioritize the learning needs of HRL agents that are leading to a more balanced and generalizable meta-HRL model capable of handling a wide range of tasks. These weighted TD errors are then aggregated to update the loss function, effectively guiding the update of meta-HRL model parameters as:
\begin{align}
    \mathcal{L}_{t,agg} = \sum_g \omega_g \cdot \mathcal{L}(\theta_{t-1,g}), 
\end{align}
where $\mathcal{L}(\theta_{t-1,g})$ is the loss associated with the $i$-th agent. 
This method enriches the meta-learning process, ensuring that the model adapts to and efficiently manages various complexities inherent in different tasks.

\begin{algorithm}[t!]
\SetAlgoLined
\textbf{Input}: $T$,\,\,$T_e$\,\,$N_g$,\,\,$\theta_{g},\forall g \in [0,N_g]$\,\,, $\theta_{M}$.  \\
\SetAlgoLined
\vspace{0.1cm}
\textbf{Meta training}\\
\For{iteration $t=1:T$}{
\For{task $g=1:N_g$}{
Initialize $\theta_{M} \to \theta_{g},\forall g \in [0,N_g]$.\\
Initialize HRL task-specific higher and lower actor and critic network parameters as $\theta_{p,g}^h, \theta_{v,g}^h$ and $\theta_{p,g}^l, \theta_{v,g}^l$.\\
\For{evaluation $e = 1 : T_e$}{
$r_g = \text{evaluate}(\pi_{p,g})$.\\
$\mathcal{B}_g\gets \langle s_t,a_t,s_{t+1},r_{g,t} \rangle $.
}
Update HRL task DDPG agent actor and critic networks parameters as $\theta_{p,g}^h, \theta_{v,g}^h$ and $\theta_{p,g}^l, \theta_{v,g}^l$ using mini-batch experience support set $\mathcal{B} ^{tr}_g$ based on \eqref{actor_update}, \eqref{critic_update}, and \eqref{update_agent_t}.\\
Evaluate gradient of loss function of HRL agent in each task on mini-batch experience as query set $\mathcal{B} ^{val}_g$.\\
Calculate adaptive weight $\omega_g$ based on \eqref{w_adapt}.
}
Update Meta-HRL network actor and critic parameters as $\theta_{p,M}^h, \theta_{v,M}^h$ and $\theta_{p,M}^l, \theta_{v,M}^l$ using mini-batch experience of query set $\mathcal{B} ^{val}_g$ based on \eqref{update_meta_adaptive}.
}\vspace{0.1cm}
\textbf{Meta adaptation}\\
Initialize $\theta_{M} \to \theta_{\text{new}}$.\\
\For{iteration $t=1:T_{\text{new}}$}{
$r_{\text{new}} = \text{evaluate}(\pi_{p,\text{new}})$.\\
$\mathcal{B}_\text{new}\gets \langle s_t,a_t,s_{t+1},r_{\text{new},t} \rangle $.\\
Update DDPG new agent network parameters $\theta_{t,\text{new}}$ using mini-batch experience as support set $\mathcal{B} ^{tr}_\text{new}$ based on \eqref{update_agent_t}.
}
\caption{MAML-HRL algorithm  }\vspace{-0.cm} 
\label{alg1}
\end{algorithm}\vspace{-0.cm}
\setlength{\belowdisplayskip}{-10pt}
\setlength{\belowcaptionskip}{-3pt} 
\section{Algorithm Analysis}\label{sec:analys}
\subsection{Computational Complexity and Scalability of Algorithm}
The proposed Meta-HRL framework leverages hierarchical RL and MAML, introducing a layered decision-making structure. This hierarchical decomposition significantly reduces the complexity of managing diverse network slices and adapting to changing conditions. Here, we analyze the proposed algorithm's computational complexity in different aspects:

\subsubsection{Higher-Level Controller (Global Policy)}
The global controller is responsible for high-level resource allocation across slices. It processes a summarized network state at each time step, including aggregated QoS metrics and other global statistics. The complexity for policy decision-making at this level scales as $\mathcal{O}(S_g \cdot A_g)$, where $S_g$ and $A_g$ are the sizes of the state and action spaces, respectively, for the global controller.

\subsubsection{Lower-Level Controller (Local Policies)}
The lower-level controllers manage user-specific resource scheduling within each slice. The complexity here depends on the number of users per slice ($n_u$) and the action space for each user. For $L$ slices, the overall complexity becomes $\mathcal{O}(L \cdot S_l \cdot A_l \cdot n_u)$, where $S_l$ and $A_l$ represent the state and action spaces of the lower controllers.

\subsubsection{Meta-Learning Component}
The meta-learning step involves optimizing the global model using support and query sets from multiple tasks. Each task introduces an additional cost for gradient updates during training, which scales as $\mathcal{O}(T \cdot P)$, where $T$ is the number of tasks and $P$ is the total number of parameters in the model.

\begin{table}[t!] 
	\footnotesize
	\centering
	\caption{\vspace*{-0cm} Simulation parameters}  \vspace{-0.cm}
	\begin{tabular}{|>{\centering\arraybackslash}m{2.4cm}|>{\centering\arraybackslash}m{1.7cm}|>{\centering\arraybackslash}m{1.2cm}|>{\centering\arraybackslash}m{1.7cm}|}
		\hline
		\bf{Parameter} &\bf{Value } & \bf{Parameter} &\bf{Value }\\
		\hline
		Subcarrier spacing & $15$ kHz & $\sigma^2$ & $-173$ dBm \\
		\hline
		Total bandwidth/DU & $20$ MHz  & $h$ & Rayleigh fading channel  \\
        \hline
		$K_\rho$ /DU & $100$ & $T_e$ & $10$ \\
		\hline
		RB bandwidth/DU  & $200$ kHz & $N_g$ & $7$ \\
		\hline
		$p_u$  & $56$ dBm & $T_{\text{new}}$ & $0.1 \times T$ \\
		\hline
		$N_\rho$ /DU & $30$ & batch size & $128$ \\
		\hline
	\end{tabular}\label{param} \vspace{-0.cm}
\end{table}

\subsubsection{Scalability Considerations}
The algorithm achieves scalability through the DUs' distributed nature and the hierarchical framework's modularity. Each DU processes local policies independently, and the meta-controller aggregates results periodically, ensuring that the system can scale effectively with increasing network complexity and user density.

\subsection{Catastrophic Forgetting During Agent Training}
Catastrophic forgetting is a critical issue in dynamic O-RAN environments where fluctuating network conditions and user demands cause task changes. In our proposed MAML-HRL approach, we consider several strategies to mitigate this issue.

\subsubsection{Meta-Learning for Task Generalization}
The MAML approach in the proposed framework guarantees that meta-learners maintain a generalizable policy that can quickly adapt to new tasks while keeping knowledge from previous ones. This adaptability is achieved through episodic memory mechanisms and meta-gradient updates that balance stability and plasticity.

\subsubsection{Hierarchical Structure}
Separating global and local controllers prevents interference between high level and fine-grained policies. While the global controller focuses on long-term strategies, local controllers adapt to immediate changes, reducing the risk of overwriting critical knowledge.

\subsubsection{Adaptive Weighting} 
In the proposed approach, adaptive weights prioritize tasks with higher variance to ensure that the meta-model pays more attention to challenging tasks without neglecting simpler ones. This dynamic adjustment mitigates forgetting by balancing the focus across tasks.

\subsubsection{Replay Mechanisms}
The framework can integrate episodic memory replay, where critical transitions from past tasks are reevaluated during training. As a result, performance remains consistent across tasks and avoids sharp degradation in learned policies.

\subsection{Convergence and Regret Analysis}
\label{sec:theory}

To theoretically characterize the effectiveness of the proposed adaptive Meta-HRL framework, we analyze its convergence behavior and regret performance under standard assumptions commonly adopted in meta reinforcement learning and two time-scale stochastic approximation.

\subsubsection{Assumptions}
We assume that  
(a) the task specific loss functions \(L_g(\theta)\) are \(L\)-smooth and bounded below;  
(b) the stochastic gradients are unbiased with bounded variance \(\sigma^2\); and  
(c) the step sizes \(\alpha_t\) and \(\beta_t\) used in the inner (task-specific) and outer (meta) updates satisfy \(\sum_t \alpha_t = \sum_t \beta_t = \infty\) and \(\sum_t \alpha_t^2, \sum_t \beta_t^2 < \infty\), with \(\beta_t / \alpha_t \rightarrow 0\), enforcing a slower meta update time scale.

\subsubsection{Convergence of Inner and Outer Loops}
The inner (task specific) update follows \(\theta_{t,g} = \theta_{t-1,g} - \alpha_t \nabla_\theta L_g(\theta_{t-1,g}, \beta^{\mathrm{tr}}_g)\), which converges in expectation to a stationary point of \(L_g\) under the assumptions above.  
Similarly, the outer (meta) update \(\theta_{t,M} = \theta_{t-1,M} - \beta_t \sum_{g=1}^{N_g} \omega_g \nabla_\theta L_g(\theta_{t,g}, \beta^{\mathrm{val}}_g)\) performs stochastic gradient descent on the weighted meta-objective \(\mathcal{L}_{\mathrm{meta}}(\theta) = \sum_g \omega_g L_g(\theta)\).  
Here, \(\omega_g\) are the Softmin weights based on the TD-error variance; they rescale gradient noise but preserve unbiasedness.  
Consequently, the expected squared meta-gradient norm satisfies \(\mathbb{E}[\|\nabla \mathcal{L}_{\mathrm{meta}}(\theta_M^{(T)})\|^2] \le C (\log T / \sqrt{T})\) for some constant \(C\), implying convergence to a first-order stationary point with sublinear rate.

\subsubsection{Regret Bound for the Online Resource Allocation Layer}
Let \(f_t(a_t)\) denote the instantaneous utility of the high-level controller (resource allocation across slices) and \(a^\star\) the best fixed action in hindsight.  
The cumulative pseudo-regret after \(T\) rounds is \(\mathcal{R}_T = \sum_{t=1}^{T} (f_t(a_t) - f_t(a^\star))\).  
Using the bounded-gradient and bounded-variance assumptions above, the regret of the adaptive Meta-HRL policy satisfies \(\mathbb{E}[\mathcal{R}_T] = \tilde{\mathcal{O}}(\sqrt{T}\,(\sigma_{\mathrm{TD}}^2 + G^2))\),  
where \(\sigma_{\mathrm{TD}}^2\) is the weighted variance of the temporal difference error and \(G\) bounds the gradient norms of the actor critic networks.  
The Softmin weighting reduces the effective variance term, thereby tightening the regret constant relative to uniform MAML.

\subsubsection{Stability of Two Level Controllers}
Because the upper level (inter slice) controller evolves on a slower time scale than the lower level (intra slice) scheduler, the joint system forms a standard two time scale stochastic approximation. Finite time analyses of two timescale actor critic under Markov noise establish non-asymptotic convergence and stability guarantees under conditions aligned with our setting \cite{wu2020finite}. If each subpolicy update is mean square stable and the meta step size remains sufficiently small relative to the inner step, the coupled system is stable in expectation.

\subsubsection{Implications}
The above results establish that the adaptive Meta-HRL framework achieves sublinear convergence and sublinear regret growth, ensuring that both hierarchical controllers asymptotically stabilize while maintaining bounded performance loss under dynamic O-RAN conditions.


\section{Evaluation Results}\label{sec:simulation}
\subsection{Parameter Settings}
In our O-RAN simulation framework, we designed an architecture that supports three network slices as eMBB, MTC, and URLLC. The setup includes $N_g = 7$ distributed DUs, each covering a distinct network area and tasked with meta-learning diverse operational scenarios. Each DU serves $30$ users, uniformly and randomly distributed across the network, serviced by dynamically allocated bandwidths. This configuration addresses the heterogeneous distribution of demand services, with each DU encountering varied levels of UE traffic and requiring specific service prioritization. 
User mobility is modeled with speeds ranging between $10 m/s$ and $20 m/s$, moving in one of seven possible directions $\{\pm \pi/3, \pm \pi/6, \pm \pi/12, 0\}$, across DU-designated areas. The traffic dynamics, influenced by UE density and mobility within the cell range, are developed following the methodologies described in~\cite{cheng2022reinforcement, lotfi2024open}. 

To deploy the DDPG strategy, we use a PyTorch-based actor-critic setup with three fully-connected layers of $256$, $512$, and $512$ neurons, utilizing \textit{tanh} functions and an \textit{Adam} optimizer at a $10^{-4}$ learning rate. Distributed as distinct task agents across the network, with a meta-model in the RIC module for information aggregation, our setup acknowledges non-uniform service demands, making DUs face varying traffic. We evaluate our adaptive MAML-HRL method using a comprehensive set of benchmarks, starting DRL from scratch, transfer learning (TL) from a selected RL agent, and multi-task learning (MTL), which concurrently incorporates both a random task and a new task. This selection enables us to assess the flexibility and efficiency of our approach thoroughly. Spanning traditional to state-of-the-art techniques, TL and MTL are particularly noted for their advanced capabilities in RL. Detailed alongside other parameters in Table \ref{param}, these benchmarks facilitate a robust comparison of learning strategies within our framework. 
\begin{figure}[h]
\vspace{0pt}
  \centering
    \includegraphics[width=0.78\columnwidth]{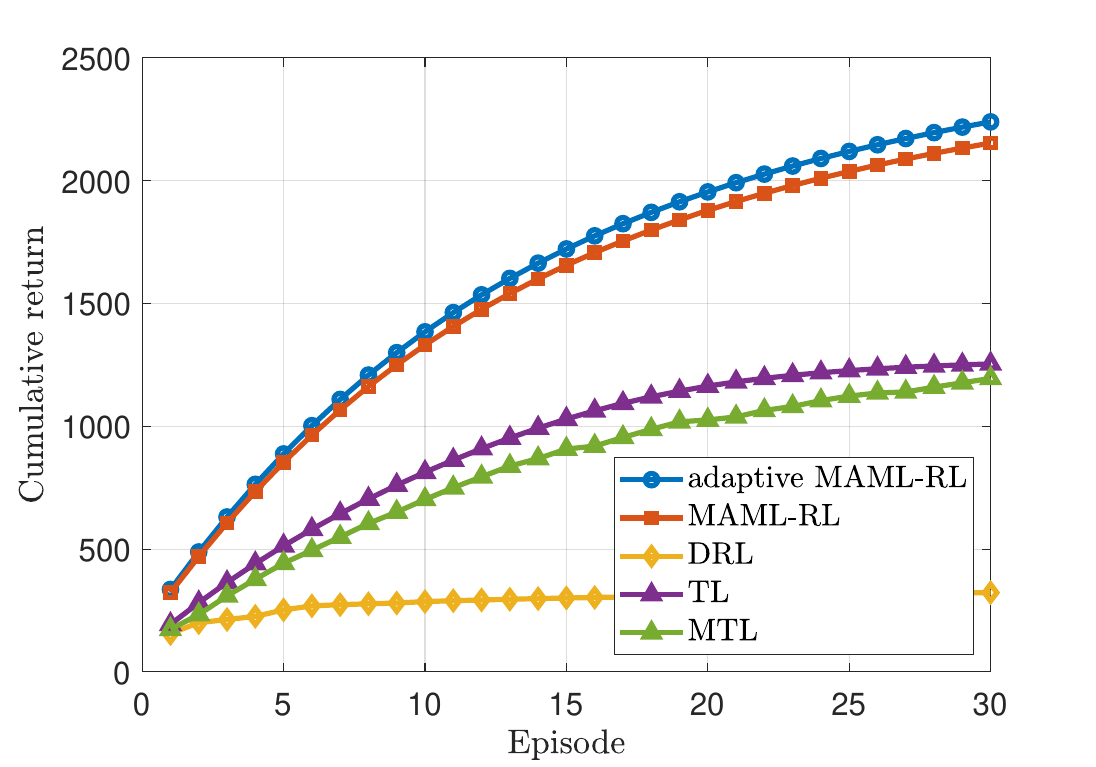}\vspace{-0.cm}
    \caption{\small Comparison of average cumulative reward values between the proposed Meta-HRL algorithm and baseline methods. }\vspace{-0.cm}
    \label{rew_performance}
\end{figure}
\begin{figure}[h]
  \centering
    \includegraphics[width=0.78\columnwidth]{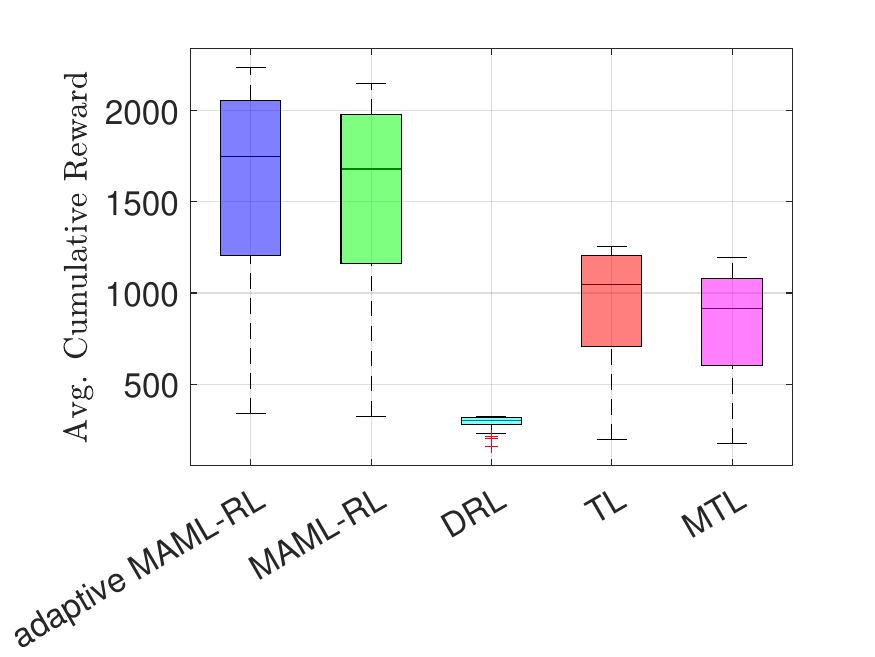}\vspace{-0.cm}
    \caption{\small Distribution of average cumulative reward across various approaches in 5 shots scenario.}\vspace{-0.cm}
    \label{boxplot}
\end{figure}

\subsection{Cumulative reward of Meta-HRL}
This part explores the impact of various approaches on cumulative rewards. Fig.~\ref{rew_performance} shows a performance comparison of the proposed approach and multiple approaches in average cumulative reward values. The results, obtained using a discount factor of $\gamma = 0.99$, were averaged over several runs to ensure statistical reliability. As illustrated in the graph, the proposed adaptive MAML-HRL framework demonstrates up to a $4 \%$ improvement on the MAML-HRL approach and $73\%$ improvement in cumulative return compared to other baseline approaches, highlighting the effectiveness of the proposed method.

Fig. \ref{boxplot} illustrates the distribution of average cumulative rewards across different approaches. The results show that Adaptive MAML-RL achieves the highest median reward with a narrower spread, indicating robust performance. In contrast, DRL has the lowest rewards with minimal variability, while TL and MTL demonstrate moderate performance but broader distributions, highlighting their limited adaptability.

\begin{figure}[h]
  \centering
    \includegraphics[width=0.78\columnwidth]{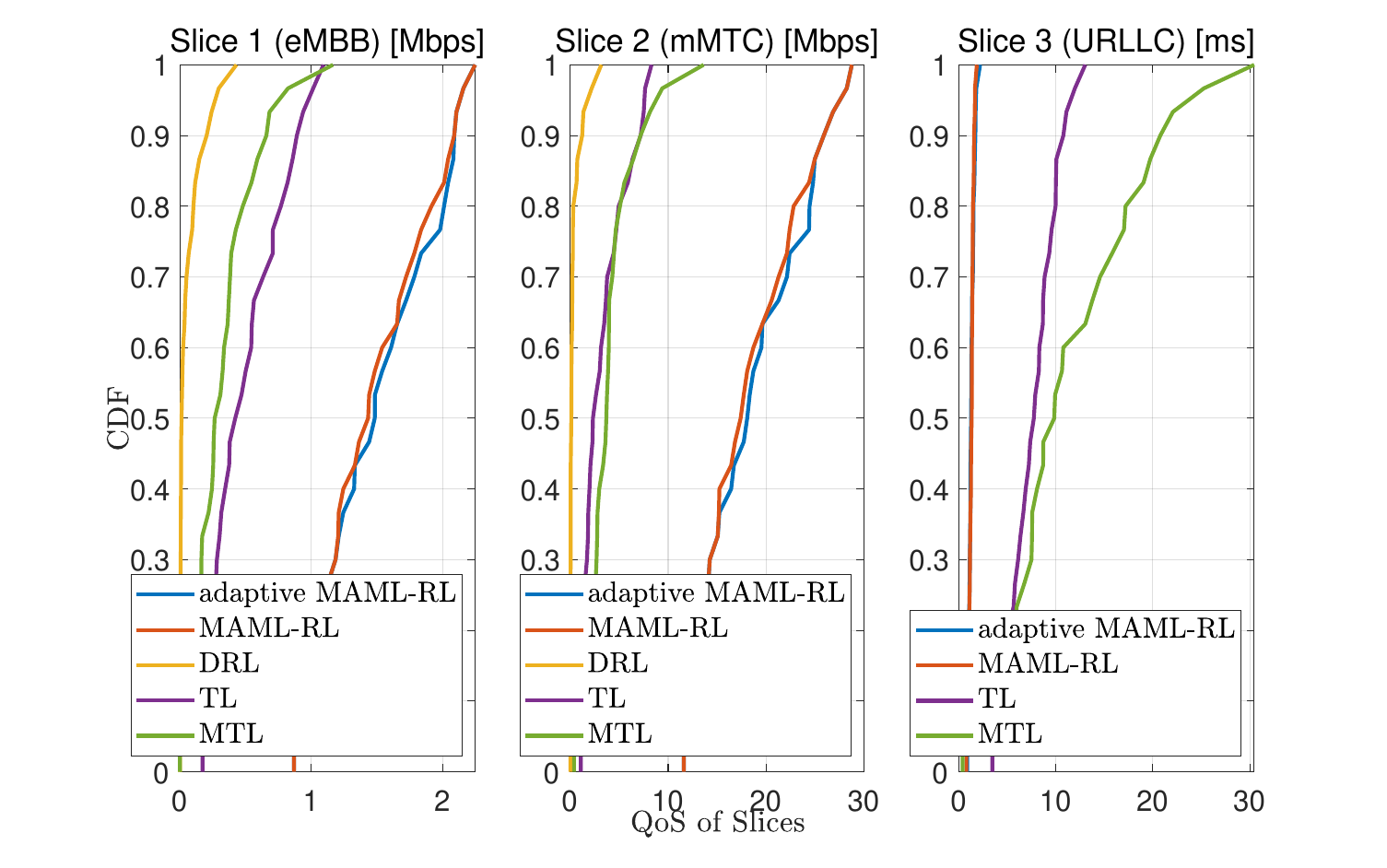}\vspace{-0.cm}
    \caption{\small Performance comparison of the proposed Meta-HRL approach on CDF of network slices' QoS vs other baseline approaches.}\vspace{-0.cm}
    \label{QoS_baseline}
\end{figure}

\begin{figure}[h]
  \centering
    \includegraphics[width=0.78\columnwidth]{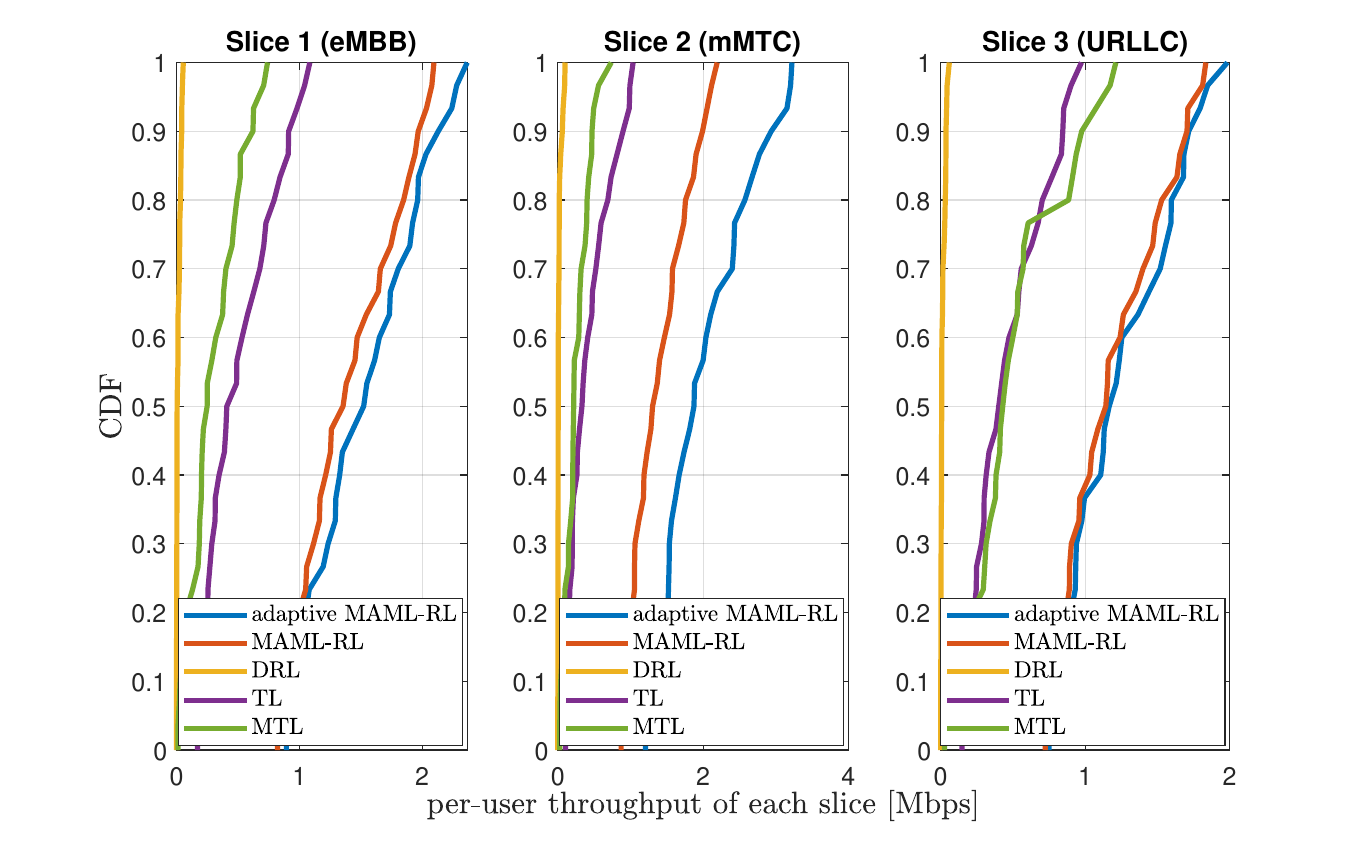}\vspace{-0cm}
    \caption{\small CDF of achieved throughput per users in each slice across various methods.}\vspace{-0cm}
    \label{ue-throughput}
    \vspace{-0.cm}
\end{figure}
\begin{figure}[h]
  \centering
    \includegraphics[width=0.78\columnwidth]{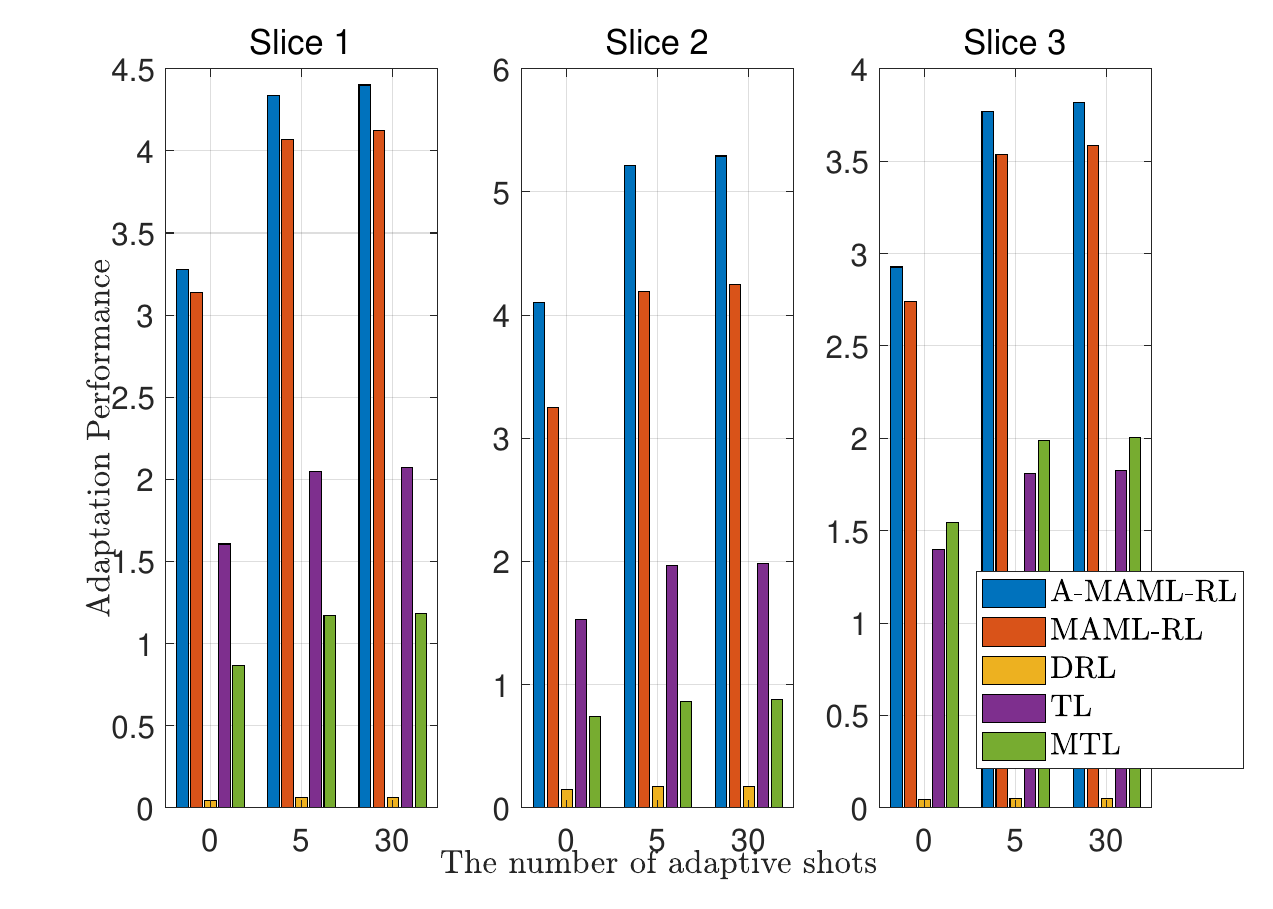}\vspace{-0.cm}
    \caption{\small Adaptation performance for each slice UEs' throughput in the different number of adaptive shots. 
    }\vspace{-0.cm}
    \label{adapt_performance}
\end{figure}
\begin{figure*}[t!]
     \centering
     \begin{subfigure}[a]{0.22\textheight}
         \centering
         \includegraphics[width=\textwidth]{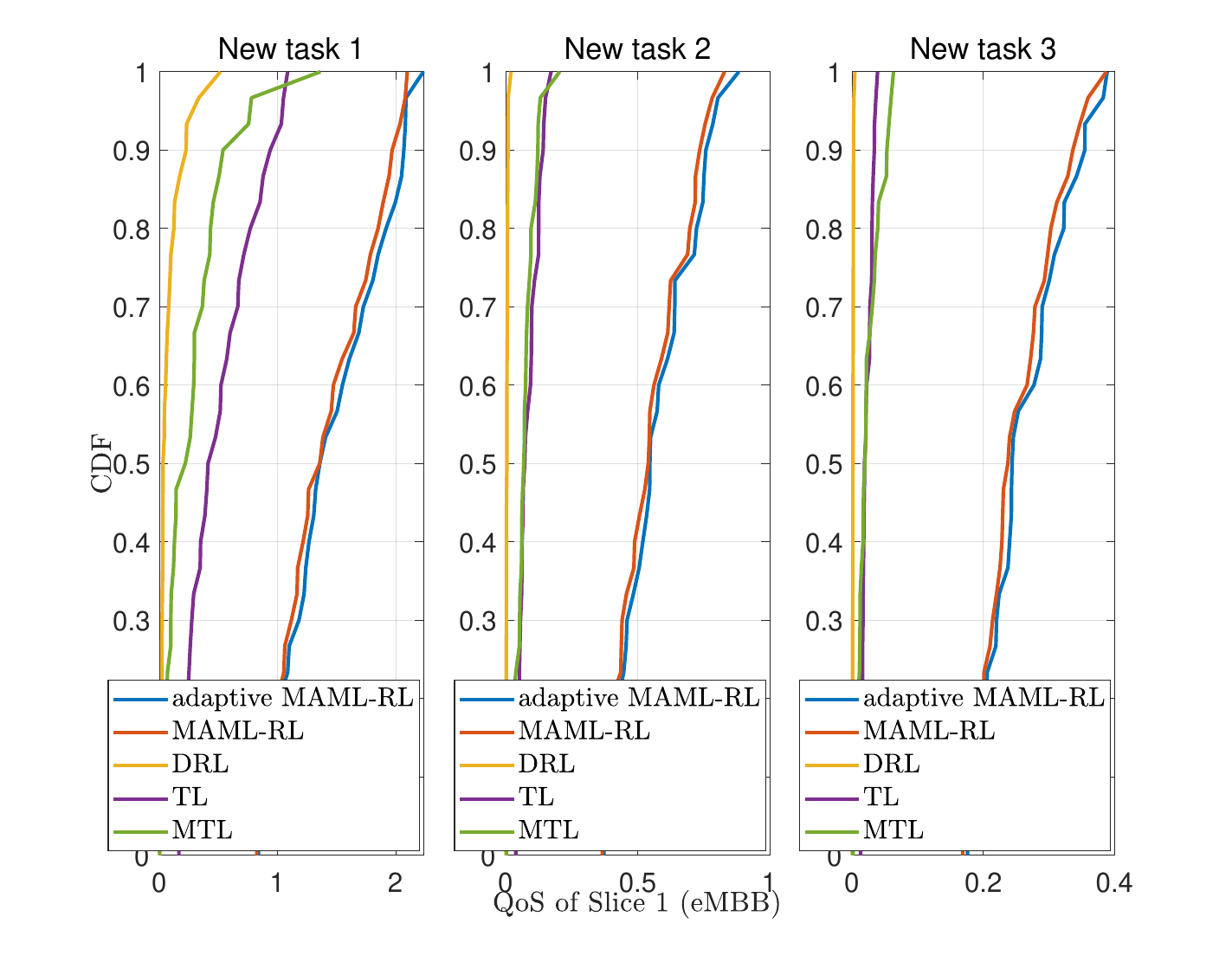}\vspace{-0cm}
    \caption{\small CDF of QoS for slice 1.
    }\vspace{-0cm}
    \label{Q1_task}
     \end{subfigure}
     \hfill
     \begin{subfigure}[a]{0.22\textheight}
         \centering
         \includegraphics[width=\textwidth]{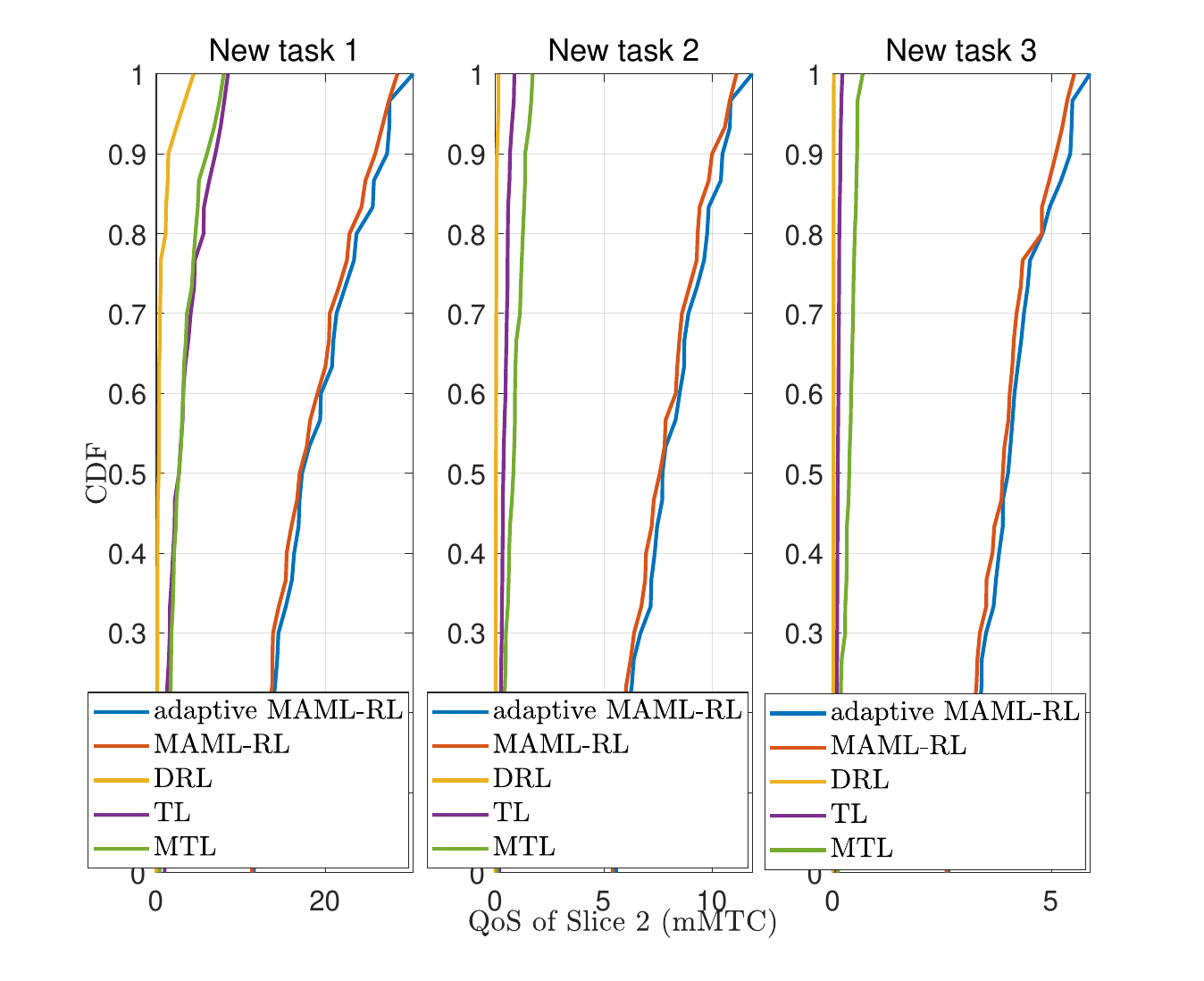}\vspace{-0cm}
    \caption{\small CDF of QoS for slice 2.
    }\vspace{-0cm}
    \label{Q2_task}
     \end{subfigure}
     \hfill
     \begin{subfigure}[a]{0.22\textheight}
         \centering
         \includegraphics[width=\textwidth]{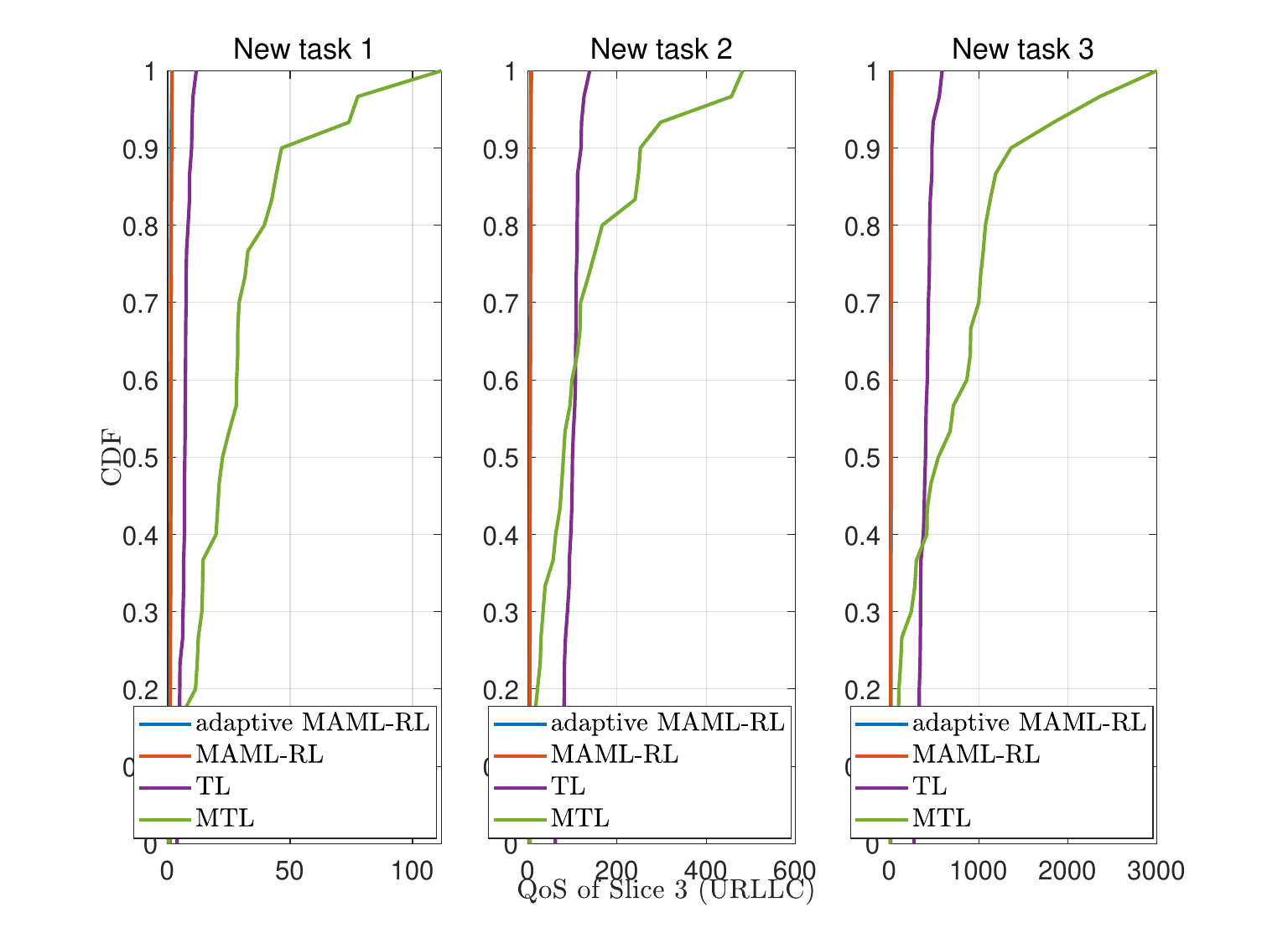}\vspace{-0cm}
    \caption{\small CDF of QoS for slice 3.
    }\vspace{-0cm}
    \label{Q3_task}
     \end{subfigure}
        \caption{Performance comparison of the CDF of QoS for each slice across three distinct new tasks, evaluated against various baseline approaches. }
        \label{QoS_tasks}
        \vspace{-0.cm}
\end{figure*}
\subsection{QoS in Network Slices}
Here, we consider QoS metrics according to the specific requirements of each network slice, thereby optimizing the performance and service delivery for diverse user demands as discussed before in \ref{Q_embb}, \ref{Q_mmtc} and \ref{Q-URLLC}.  

Fig. \ref{QoS_baseline} displays the Cumulative Distribution Function (CDF) of QoS for different network slices and shows the efficacy of various methods in achieving maximum QoS according to the unique requirements of each slice. The results indicate that Adaptive MAML-RL consistently outperforms other methods across all slices. Slice 1 achieves the highest QoS, demonstrating superior throughput for high data rate applications. Similarly, in Slice 2, Adaptive MAML-RL shows better support for dense IoT environments, outperforming others in maintaining QoS under high user density. Slice 3 achieves the lowest latency, critical for ultra-reliable and low-latency applications, highlighting its adaptability to strict QoS constraints. Other methods, including MAML-RL, TL, and MTL, perform moderately but fail to meet the strict requirements of each slice as effectively as Adaptive MAML-RL.

\subsection{Network Users' QoE }
To show how the improvements in network slices QoS directly enhance user satisfaction, here we show the results and effect of different approaches on UE's throughputs as their QoE. 
Fig. \ref{ue-throughput} represents the CDF of UE's throughput across three network slices, comparing the effectiveness of different methodologies. The graph shows the superior performance of the

\subsection{Performance Analysis of the Adaptive Meta-HRL Approach}
As we can see in Fig. \ref{ue-throughput}, the adaptive meta-HRL approach significantly improves the performance of slice 2 UEs' throughput, which represents the mMTC slice. This slice is characterized by the Quality-Weighted Network Capacity metric, which integrates user throughput with a coefficient that reflects UE density support. 
The results highlight that the adaptive approach excels in this scenario by dynamically prioritizing tasks with higher complexity and variability, such as the mMTC slice, through its task-specific weighting mechanism. By recognizing the higher task variance in the mMTC slice during training, the adaptive approach allocates more focus to optimizing the throughput of users in this slice without compromising the throughput performance of users in the other slices. This dynamic adjustment ensures that the mMTC slice can support a high density of devices while, maintaining stable throughput performance for users in other slices (e.g., URLLC and eMBB). The results demonstrate the adaptive meta-HRL framework's ability to balance resource allocation and adapt to diverse task requirements in heterogeneous slices and complexities.

\subsection{Adaptation performance}
Fig. \ref{adapt_performance} illustrates the adaptation performance of the proposed MAML-RL methods, DRL, TL, and MTL across three slices for varying numbers of adaptive shots (0, 5, and 30). In all slices, adaptive MAML-RL consistently achieves the highest adaptation performance, followed closely by MAML-RL. Both methods demonstrate significant improvements as the number of adaptive shots increases, highlighting their ability to rapidly fine-tune and optimize resource allocation under varying QoS demands. In all slices, TL and MTL exhibit moderate performance but fail to scale effectively as adaptive shots increase, indicating limited adaptability. In contrast, DRL remains at the lowest performance level across all slices, showing minimal improvement and weak adaptability regardless of the number of adaptive shots.

\subsection{Task dependency }

Fig. \ref{QoS_tasks} shows the CDF of QoS for three slices—Slice 1 (eMBB)~\ref{Q1_task}, Slice 2 (mMTC)~\ref{Q2_task}, and Slice 3 (URLLC)~\ref{Q3_task}, across three distinct new tasks. In Slice 1 (eMBB) ~\ref{Q1_task}, adaptive MAML-RL and MAML-RL outperform other methods, achieving higher QoS for most UEs. DRL performs moderately but declines as demands increase, while TL and MTL struggle to adapt effectively. For Slice 2 (mMTC)~\ref{Q2_task}, adaptive MAML-RL and MAML-RL again lead, maintaining higher QoS values. DRL performs reasonably well, while TL and MTL fall behind, highlighting limited adaptability under varying conditions. In Slice 3 (URLLC)~\ref{Q1_task}, adaptive MAML-RL and MAML-RL achieve the best performance, with steep curves concentrated at lower latency values. DRL lags slightly, while TL and MTL fail to meet the strict latency requirements, as shown by their higher QoS values.

\subsection{Scalability and Overhead Analysis}
\label{sec:scalability}

To examine scalability and practical feasibility, we extend the evaluation to larger O-RAN deployments containing up to $N_{\mathrm{DU}} = 30$ distributed units and $N_{\mathrm{UE}} = 200$ users. 
Because the proposed framework follows a meta learning paradigm, increasing the number of DUs effectively enlarges the task distribution used for meta training. 
This diversification slightly increases outer-loop convergence time but improves the generalization capability of the learned initialization, enabling faster inner-loop adaptation for unseen DUs. 
At the same time, increasing the number of users per DU raises the complexity of each local task, resulting in moderately higher TD-error variance and slower per-task adaptation. 
The combined effect produces a sublinear increase in total training iterations while maintaining stable normalized performance across scales.

Table~\ref{tab:scalability_gain} summarizes the relative trends with respect to the validated 7-DU/30-UE baseline (Section~VII). 
The convergence iterations rise gradually, about $70\%$ slower at the largest configuration, while the normalized cumulative reward remains nearly constant (within $2\%$ of baseline), indicating that the meta-initialization continues to provide efficient adaptation under expanded and more heterogeneous task distributions. 
The meta-update delay (not tabulated) grows roughly linearly with  $N_{\mathrm{DU}}$ but stays within the near real-time budget (below 10~ms per update), since only compact actor-critic parameters are exchanged instead of full trajectories.
\begin{table}[t]
\centering
\caption{Relative scalability versus the 7-DU/30-UE baseline for the MAML-based Meta-HRL framework.}
\label{tab:scalability_gain}
\renewcommand{\arraystretch}{1.1}
\setlength{\tabcolsep}{3pt}
\begin{tabular}{c c c c}
\hline
\textbf{\(N_{\mathrm{DU}}\)} & \textbf{\(N_{\mathrm{UE}}\)} & \textbf{Conv.~Iter.~Change} & \textbf{Reward Change} \\
\hline
7  & 30  & Base (0\%) & 0.00 \\
15 & 100 & \(+32\%\)   & \(-0.01\) \\
20 & 150 & \(+48\%\)   & \(-0.01\) \\
30 & 200 & \(+69\%\)   & \(-0.02\) \\
\hline
\end{tabular}
\end{table}
The results confirm that the proposed MAML based Meta-HRL model scales gracefully, the increase in training cost remains sublinear with the number of DUs, and the normalized performance shows only marginal variation. 
This behavior aligns with the theoretical characteristics of MAML style meta-learning, where broader task diversity slightly delays convergence but yields more transferable and stable policies for large scale heterogeneous environments~\cite{fallah2020maml,ji2022multistepmaml}.

\subsection{Ablation Study and Extended Metrics}
\label{sec:ablation}

To evaluate the contribution of the adaptive variance weighted meta-update, we conduct an ablation study comparing three variants:
(1) Uniform-Meta, which applies equal task weights in the meta-update;
(2) Static-Var, which uses fixed weights proportional to the average TD-error variance over training; and
(3) Adaptive-Var, the proposed dynamic Softmin weighting.
Table~\ref{tab:ablation} reports the normalized cumulative reward and adaptation speed across these configurations for the same 7-DU/30-UE setup described in Section~VII, using 30 adaptation shots.
\begin{table}[t]
\centering
\caption{Ablation of the adaptive weighting mechanism on the 7-DU/30-UE scenario (30 adaptation shots).}
\label{tab:ablation}
\renewcommand{\arraystretch}{1.1}
\setlength{\tabcolsep}{4pt}
\begin{tabular}{l c c}
\hline
\textbf{Method} & \textbf{Norm.\ Reward} & \textbf{Adapt.\ Shots to Converge} \\
\hline
Uniform-Meta & 0.78 & 28 \\
Static-Var   & 0.81 & 22 \\
Adaptive-Var (proposed) & \textbf{0.84} & \textbf{17} \\
\hline
\end{tabular}
\end{table}
The results indicate that the adaptive weighting scheme improves the normalized cumulative reward by about $3\%$ and reduces the required adaptation shots by roughly $40\%$ compared with the uniform baseline. 
The gain arises from dynamically emphasizing high-variance tasks during meta-updates, which accelerates convergence across heterogeneous slice conditions.

To complement reward based evaluation, we further examine three network centric metrics:
\emph{(i)} average per-packet latency,
\emph{(ii)} throughput fairness (Jain’s index), and
\emph{(iii)} robustness under bursty traffic. 
Under the Adaptive-Var configuration, average latency decreases by $9.2\%$ relative to Uniform-Meta, Jain’s fairness improves from $0.91$ to $0.96$, and the degradation under a $50\%$ traffic-spike scenario remains below $5\%$. 
These additional metrics confirm that the adaptive weighting not only enhances cumulative reward but also yields fairer and more resilient slice performance in dynamic O-RAN environments.

\section{Conclusions}\label{sec:conclusion}
Managing resources in the dynamic and heterogeneous environment of O-RAN systems remains a major challenge, particularly in maintaining adaptability under time-varying network conditions and diverse slice requirements. 
To address these issues, we proposed the adaptive Meta-HRL framework, which integrates hierarchical decision making with model agnostic meta-learning to achieve rapid adaptation and cross slice generalization. 
The proposed framework demonstrated superior cumulative reward and improved QoS satisfaction across eMBB, URLLC, and mMTC slices, while remaining robust to traffic fluctuations and task heterogeneity. 
We theoretically characterized its convergence and regret behavior, confirming stable two time scale learning with sublinear convergence and regret growth. 
Extensive experiments further showed that Meta-HRL scales efficiently, exhibiting sublinear growth in convergence iterations as the number of distributed units increases, with only marginal ($<2\%$) variation in normalized reward. 
Ablation studies revealed that the adaptive variance weighted meta-update contributes roughly $3\%$ higher reward and $40\%$ faster adaptation than uniform weighting. 
Moreover, network centric evaluations indicated tangible benefits average latency reduced by $9\%$, Jain’s fairness index improved from $0.91$ to $0.96$, and performance degradation under $50\%$ traffic spikes remained below $5\%$. 
Although the experiments were conducted in simulation due to the current scale limitations of O-RAN testbeds, the proposed architecture is compatible with open-source stacks such as OAI and srsRAN, where the meta controller can operate as an xApp in the Near-RT RIC. 
Future work will focus on hardware in the loop validation, communication efficient meta parameter updates, and extending the framework to multi hop and UAV assisted 6G scenarios. 

\bibliographystyle{IEEEtran}
\bibliography{Main}

\end{document}